">INTERNATIONAL COMPUTER SCIENCE INSTITUTE 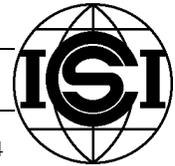

1947 Center St. • Suite 600 • Berkeley, California 94704-1198 • (510) 643-9153 • FAX (510) 643-7684# An Efficient Probabilistic Context-Free Parsing Algorithm that Computes Prefix Probabilities

">Andreas Stolcke[*]">TR-93-065

November 1993
(Revised November 1994)## Abstract

abstract">We describe an extension of Earley's parser for stochastic context-free grammars that computes the following quantities given a stochastic context-free grammar and an input string: a) probabilities of successive prefixes being generated by the grammar; b) probabilities of substrings being generated by the nonterminals, including the entire string being generated by the grammar; c) most likely (Viterbi) parse of the string; d) posterior expected number of applications of each grammar production, as required for reestimating rule probabilities. (a) and (b) are computed incrementally in a single left-to-right pass over the input. Our algorithm compares favorably to standard bottom-up parsing methods for SCFGs in that it works efficiently on sparse grammars by making use of Earley's top-down control structure. It can process any context-free rule format without conversion to some normal form, and combines computations for (a) through (d) in a single algorithm. Finally, the algorithm has simple extensions for processing partially bracketed inputs, and for finding partial parses and their likelihoods on ungrammatical inputs.

Abridged version to appear in *Computational Linguistics*.">[*]Present address: Speech Technology and Research Laboratory, SRI International, 333 Ravenswood Ave., Menlo Park, CA 94025, e-mail stolcke@speech.sri.com.">cmp-lg/9411029  28 Nov 1994

# Contents

# 1  Introduction

Context-free grammars are widely used as models of natural language syntax. In their probabilistic version, which defines a language as a probability distribution over strings, they have been used in a variety of applications: for the selection of parses for ambiguous inputs (Fujisaki *et al.* 1991); to guide the rule choice efficiently during parsing (Jones & Eisner 1992); to compute island probabilities for non-linear parsing (Corazza *et al.* 1991). In speech recognition, probabilistic context-free grammars play a central role in integrating low-level word models with higher-level language models (Ney 1992), as well as in non-finite state acoustic and phonotactic modeling (Lari & Young 1991). In some work, context-free grammars are combined with scoring functions that are not strictly probabilistic (Nakagawa 1987), or they are used with context-sensitive and/or semantic probabilities (Magerman & Marcus 1991; Magerman & Weir 1992; Jones & Eisner 1992; Briscoe & Carroll 1993).

Although clearly not a perfect model of natural language, stochastic context-free grammars (SCFGs) are superior to non-probabilistic CFGs, with probability theory providing a sound theoretical basis for ranking and pruning of parses, as well as for integration with models for non-syntactic aspects of language. All of the applications listed above involve (or could potentially make use of) one or more of the following standard tasks, compiled by Jelinek & Lafferty (1991).[1]

1. What is the probability that a given string $x$ is generated by a grammar $G$?

2. What is the single most likely parse (or derivation) for $x$?

3. What is the probability that $x$ occurs as a prefix of some string generated by $G$ (the *prefix probability* of $x$)?

4. How should the parameters (e.g., rule probabilities) in $G$ be chosen to maximize the probability over a training set of strings?

The algorithm described in this paper can compute solutions to all four of these problems in a single framework, with a number of additional advantages over previously presented isolated solutions.

Most probabilistic parsers are based on a generalization of bottom-up chart parsing, such as the CYK algorithm. Partial parses are assembled just as in non-probabilistic parsing (modulo possible pruning based on probabilities), while substring probabilities (also known as "inside" probabilities) can be computed in a straightforward way. Thus, the CYK chart parser underlies the standard solutions to problems (1) and (4) (Baker 1979), as well as (2) (Jelinek 1985). While the Jelinek & Lafferty (1991) solution to problem (3) is not a direct extension of CYK parsing they nevertheless present their algorithm in terms of its similarities to the computation of inside probabilities.

In our algorithm, computations for tasks (1) and (3) proceed incrementally, as the parser scans its input from left to right; in particular, prefix probabilities are available as soon as the prefix has been seen, and are updated incrementally as it is extended. Tasks (2) and (4) require one more (reverse) pass over the chart constructed from the input.

Incremental, left-to-right computation of prefix probabilities is particularly important since that is a necessary condition for using SCFGs as a replacement for finite-state language models in many applications, such a speech decoding. As pointed out by Jelinek & Lafferty (1991), knowing probabilities $P(x_0 \ldots x_i)$ for arbitrary prefixes $x_0 \ldots x_i$ enables probabilistic prediction of possible follow-words $x_{i+1}$, as $P(x_{i+1}|x_0 \ldots x_i) = P(x_0 \ldots x_i x_{i+1})/P(x_0 \ldots x_i)$. These conditional probabilities can then be used as word transition probabilities in a Viterbi-style decoder or to incrementally compute the cost function for a stack decoder (Bahl *et al.* 1983).

---

[1] Their paper phrases these problem in terms of context-free probabilistic grammars, but they generalize in obvious ways to other classes of models.

Another application where prefix probabilities play a central role is the extraction of $n$-gram probabilities from SCFGs (Stolcke & Segal 1994). Here, too, efficient incremental computation saves time since the work for common prefix strings can be shared.

The key to most of the features of our algorithm is that it is based on the top-down parsing method for non-probabilistic CFGs developed by Earley (1970). Earley's algorithm is appealing because it runs with best-known complexity on a number of special classes of grammars. In particular, Earley parsing is more efficient than the bottom-up methods in cases where top-down prediction can rule out potential parses of substrings. The worst-case computational expense of the algorithm (either for the complete input, or the incrementally for each new word) is as good as that of the other known specialized algorithms, but can be substantially better on well-known grammar classes.

Earley's parser (and hence ours) also deals with any context-free rule format in a seamless way, without requiring conversions to Chomsky Normal Form (CNF), as is often assumed. Another advantage is that our probabilistic Earley parser has been extended to take advantage of partially bracketed input, and to return partial parses on ungrammatical input. The latter extension removes one of the common objections against top-down, predictive (as opposed to bottom-up) parsing approaches (Magerman & Weir 1992).

## 2  Overview

The remainder of the paper proceeds as follows. Section 3 briefly reviews the workings of an Earley parser without regard to probabilities. Section 4 describes how the parser needs to be extended to compute sentence and prefix probabilities. Section 5 deals with further modifications for solving the Viterbi and training tasks, for processing partially bracketed inputs, and for finding partial parses. Section 6 discusses miscellaneous issues and relates our work to the literature on the subject. In Section 7 we summarize and draw some conclusions.

To get an overall idea of probabilistic Earley parsing it should be sufficient to read Sections 3, 4.2 and 4.4. Section 4.5 deals with a crucial technicality, and later sections mostly fill in details and add optional features.

We assume the reader is familiar with the basics of context-free grammar theory, such as given in Aho & Ullman (1972:chapter 2). Some prior familiarity with probabilistic context-free grammars will also be helpful. Jelinek *et al.* (1992) provide a tutorial introduction covering the standard algorithms for the four tasks mentioned in the introduction.

**Notation**  The input string is denoted by $x$. $|x|$ is the length of $x$. Individual input symbols are identified by indices starting at 0: $x_0, x_1, \ldots, x_{|x|-1}$. The input alphabet is denoted by $\Sigma$. Substrings are identified by beginning and end positions $x_{i \ldots j}$. The variables $i, j, k$ are reserved for integers referring to positions in input strings. Latin capital letters $X, Y, Z$ denote nonterminal symbols. Latin lowercase letters $a, b, \ldots$ are used for terminal symbols. Strings of mixed nonterminal and terminal symbols are written using lowercase Greek letters $\lambda, \mu, \nu$. The empty string is denoted by $\epsilon$.

## 3  Earley Parsing

An Earley parser is essentially a generator that builds left-most derivations of strings, using a given set of context-free productions. The parsing functionality arises because the generator keeps track of all possible derivations that are consistent with the input string up to a certain point. As more and more of the input is revealed the set of possible derivations (each of which corresponds to a parse) can either expand as new choices are introduced, or shrink as a result of resolved ambiguities. In describing the parser it is thus appropriate and convenient to use generation terminology.

The parser keeps a set of *states* for each position in the input, describing all pending derivations.[2] These state sets together form the Earley *chart*. A state is of the form

$$i: \quad _kX \to \lambda.\mu,$$

where $X$ is a nonterminal of the grammar, $\lambda$ and $\mu$ are strings of nonterminals and/or terminals, and $i$ and $k$ are indices into the input string. States are derived from productions in the grammar. The above state is derived from a corresponding production

$$X \to \lambda\mu$$

with the following semantics:

- The current position in the input is $i$, i.e., $x_0 \ldots x_{i-1}$ have been processed so far.[3] The states describing the parser state at position $i$ are collectively called *state set i*. Note that there is one more state set than input symbols: set 0 describes the parser state before any input is processed, while set $|x|$ contains the states after all input symbols have been processed.

- Nonterminal $X$ was expanded starting at position $k$ in the input, i.e., $X$ generates some substring starting at position $k$.

- The expansion of $X$ proceeded using the production $X \to \lambda\mu$, and has expanded the right-hand side (RHS) $\lambda\mu$ up to the position indicated by the dot. The dot thus refers to the current position $i$.

A state with the dot to the right of the entire RHS is called a *complete* state, since it indicates that the left-hand side (LHS) nonterminal has been fully expanded.

Our description of Earley parsing omits an optional feature of Earley states, the *lookahead string*. Earley's algorithm allows for an adjustable amount of lookahead during parsing, in order to process LR($k$) grammars deterministically (and obtain the same computational complexity as specialized LR($k$) parsers where possible). The addition of lookahead is orthogonal to our extension to probabilistic grammars, so we will not include it here.

The operation of the parser is defined in terms of three operations that consult the current set of states and the current input symbol, and add new states to the chart. This is strongly suggestive of *state transitions* in finite-state models of language, parsing, etc. This analogy will be explored further in the probabilistic formulation later on.

The three types of transitions operate as follows.

*Prediction* For each state

$$i: \quad _kX \to \lambda.Y\mu,$$

where $Y$ is a nonterminal anywhere in the RHS, and for all rules $Y \to \nu$ expanding $Y$, add states

$$i: \quad _iY \to .\nu \quad .$$

A state produced by prediction is called a *predicted state*. Each prediction corresponds to a potential expansion of a nonterminal in a left-most derivation.

*Scanning* For each state

$$i: \quad _kX \to \lambda.a\mu,$$

where $a$ is a terminal symbol that matches the current input $x_i$, add the state

$$i+1: \quad _kX \to \lambda a.\mu$$

(move the dot over the current symbol). A state produced by scanning is called a *scanned state*. Scanning ensures that the terminals produced in a derivation match the input string.

---

[2] Earley states are also known as *items* in LR parsing, see Aho & Ullman (1972:section 5.2) and Section 6.3.
[3] This index is implicit in Earley (1970). We include it here for clarity.

*Completion* For each complete state

$$i: \quad _jY \to \nu.$$

and each state in set $j$, $j < i$, that has $Y$ to the right of the dot,

$$j: \quad _kX \to \lambda.Y\mu \quad ,$$

add the state

$$i: \quad _kX \to \lambda Y.\mu$$

(move the dot over the current nonterminal). A state produced by completion is called a *completed state*.[4] Each completion corresponds to the end of a nonterminal expansion started by a matching prediction step.

For each input symbol and corresponding state set, an Earley parser performs all three operations exhaustively, i.e., until no new states are generated. One crucial insight into the working of the algorithm is that, although both prediction and completion feed themselves, there are only a finite number of states that can possibly be produced. Therefore recursive prediction and completion at each position have to terminate eventually, and the parser can proceed to the next input via scanning.

To complete the description we need only specify the initial and final states. The parser starts out with

$$0: \quad _0 \to .S,$$

where $S$ is the sentence nonterminal (note the empty left-hand side). After processing the last symbol, the parser verifies that

$$l: \quad _0 \to S.$$

has been produced (among possibly others), where $l$ is the length of the input $x$. If at any intermediate stage a state set remains empty (because no states from the previous stage permit scanning) the parse can be aborted because an impossible prefix has been detected.

States with empty LHS such as those above are useful in other contexts, as will be shown in Section 5.4. We will collectively refer to them as *dummy states*. Dummy states enter the chart only as a result of initialization, as opposed to being derived from grammar productions.

It is easy to see that Earley parser operations are *correct*, in the sense that each chain of transitions (predictions, scanning steps, completions) corresponds to a possible (partial) derivation. Intuitively, it is also true that a parser that performs these transitions exhaustively is *complete*, i.e., it finds all possible derivations. Formal proofs of these properties are given in the literature, e.g., Aho & Ullman (1972). The relationship between Earley transitions and derivations will be stated more formally in the next section.

The parse trees for sentences can be reconstructed from the chart contents. We will illustrate this in Section 5 when discussing Viterbi parses.

Table 1 shows a simple grammar and a trace of Earley parser operation on a sample sentence.

Earley's parser can deal with any type of context-free rule format, even with null or $\epsilon$-productions, i.e., those that replace a nonterminal with the empty string. Such productions do however require special attention, and make the algorithm and its description more complicated than otherwise necessary. In the following sections we assume that no null productions have to be dealt with, and then summarize the necessary changes in Section 4.7. One might chose to simply preprocess the grammar to eliminate null productions, a process which is also described.

---

[4] Note the difference between "complete" and "completed" states: Complete states (those with the dot to the right of the entire RHS) are the result of a completion or scanning step, but completion also produces states which are not yet complete.

(a)

| | | | | | |
|---|---|---|---|---|---|
| | S | $\to$ | NP VP | Det $\to$ | a |
| | NP | $\to$ | Det N | N $\to$ | circle\|square\|triangle |
| | VP | $\to$ | VT NP | VT $\to$ | touches |
| | VP | $\to$ | VI PP | VI $\to$ | is |
| | PP | $\to$ | P NP | P $\to$ | above\|below |

(b)

| | a | circle | touches | a | square |
|---|---|---|---|---|---|
| $_0 \to$ .S | *scanned* | *scanned* | *scanned* | *scanned* | *scanned* |
| *predicted* | $_0$Det $\to$ a. | $_1$N $\to$ circle. | $_2$VT $\to$ touches. | $_3$Det $\to$ a. | $_4$N $\to$ triangle. |
| $_0$S $\to$ .NP VP | *completed* | *completed* | *completed* | *completed* | *completed* |
| $_0$NP $\to$ .Det N | $_0$NP $\to$ Det.N | $_0$NP $\to$ Det N. | $_2$VP $\to$ VT.NP | $_3$NP $\to$ Det.N | $_4$NP $\to$ Det N. |
| $_0$Det $\to$ .a | *predicted* | $_0$S $\to$ NP.VP | *predicted* | *predicted* | $_3$VP $\to$ VT NP. |
| | $_1$N $\to$ .circle | *predicted* | $_3$NP $\to$ .Det N | $_5$N $\to$ .circle | $_0$S $\to$ NP VP. |
| | $_1$N $\to$ .square | $_2$VP $\to$ .VT NP | $_3$Det $\to$ .a | $_4$N $\to$ .square | $_0 \to$ S. |
| | $_1$N $\to$ .triangle | $_2$VP $\to$ .VI PP | | $_4$N $\to$ .triangle | |
| | | $_2$VT $\to$ .touches | | | |
| | | $_2$VI $\to$ .is | | | |
| State set 0 | 1 | 2 | 3 | 4 | 5 |

Table 1: (a) Example grammar for a tiny fragment of English. (b) Earley parser processing the sentence *a circle touches a triangle*.

## 4 Probabilistic Earley Parsing

### 4.1 Stochastic context-free grammars

A stochastic context-free grammar (SCFG) extends the standard context-free formalism by adding probabilities to each production:

$$X \to \lambda \quad [p],$$

where the rule probability $p$ is usually written as $P(X \to \lambda)$. This notation to some extent hides the fact that $p$ is a conditional probability, of production $X \to \lambda$ being chosen, given that $X$ is up for expansion. The probabilities of all rules with the same nonterminal $X$ on the LHS must therefore sum to unity. Context-freeness in a probabilistic setting translates into conditional independence of rule choices. As a result, complete derivations have joint probabilities that are simply the products of the rule probabilities involved.

The probabilities of interest mentioned in Section 1 can now be defined formally.

**Definition 1** The following quantities are defined relative to a SCFG $G$, a nonterminal $X$, and a string $x$ over the alphabet of $G$.

a) The *probability of a (partial) derivation* $\nu_1 \Rightarrow \nu_2 \Rightarrow \ldots \nu_k$ is inductively defined by

1) $P(\nu_1) = 1$
2) $P(\nu_1 \Rightarrow \ldots \Rightarrow \nu_k) = P(X \to \lambda) P(\nu_2 \Rightarrow \ldots \Rightarrow \nu_k)$,

where $\nu_1, \nu_2, \ldots, \nu_k$ are strings of terminals and nonterminals, $X \to \lambda$ is a production of $G$, and $\nu_2$ is derived from $\nu_1$ by replacing one occurrence of $X$ with $\lambda$.

b) The *string probability* $P(X \stackrel{*}{\Rightarrow} x)$ (of $x$ given $X$) is the sum of the probabilities of all left-most derivations $X \Rightarrow \ldots \Rightarrow x$ producing $x$ from $X$.[5]

c) The *sentence probability* $P(S \stackrel{*}{\Rightarrow} x)$ (of $x$ given $G$) is the string probability given the start symbol $S$ of $G$. By definition, this is also the probability $P(x|G)$ assigned to $x$ by the grammar $G$.

d) The *prefix probability* $P(S \stackrel{*}{\Rightarrow}_L x)$ (of $x$ given $G$) is the sum of the probabilities of all sentence strings having $x$ as a prefix,

$$P(S \stackrel{*}{\Rightarrow}_L x) = \sum_{y \in \Sigma^*} P(S \stackrel{*}{\Rightarrow} xy) \quad .$$

(In particular, $P(S \stackrel{*}{\Rightarrow}_L \epsilon) = 1$).

In the following, we assume that the probabilities in a SCFG are *proper* and *consistent* as defined in Booth & Thompson (1973), and that the grammar contains no *useless nonterminals* (ones that can never appear in a derivation). These restrictions ensure that all nonterminals define probability measures over strings, i.e., $P(X \stackrel{*}{\Rightarrow} x)$ is a proper distribution over $x$ for all $X$. Formal definitions of these conditions are given in Appendix A.

## 4.2 Earley paths and their probabilities

In order to define the probabilities associated with parser operation on a SCFG, we need the concept of a path, or partial derivation, executed by the Earley parser.

**Definition 2** a) An *(unconstrained) Earley path*, or simply *path*, is a sequence of Earley states linked by prediction, scanning, or completion. For the purpose of this definition, we allow scanning to operate in "generation mode," i.e., all states with terminals to the right of the dot can be scanned, not just those matching the input. (For completed states, the predecessor state is defined to be the complete state from the same state set contributing to the completion.)

b) A path is said to be *constrained* by, or *generate* a string $x$ if the terminals immediately to the left of the dot in all scanned states, in sequence, form the string $x$.

c) A path is *complete* if the last state on it matches the first, except that the dot has moved to the end of the RHS.

d) We say that a path *starts with* nonterminal $X$ if the first state on it is a predicted state with $X$ on the LHS.

e) The *length* of a path is defined as the number of *scanned* states on it.

Note that the definition of path length is somewhat counter-intuitive, but is motivated by the fact that only scanned states correspond directly to input symbols. Thus, the length of a path is always the same as the length of the input string it generates.

A constrained path starting with the initial state contains a sequence of states from state set 0 derived by repeated prediction, followed by a single state from set 1 produced by scanning the first symbol, followed by a sequence of states produced by completion, followed by a sequence of predicted states, followed by a state scanning the second symbol, and so on. The significance of Earley paths is that they are in a one-to-one correspondence with left-most derivations. This will allow us to talk about probabilities of derivations, strings and prefixes in terms of the actions performed by Earley's parser. From now on, we will use "derivation" to imply a left-most derivation.

---

[5] In a *left-most derivation* each step replaces the nonterminal furthest to the left in the partially expanded string. The order of expansion is actually irrelevant for this definition, due to the multiplicative combination of production probabilities. We restrict summation to left-most derivations to avoid counting duplicates, and because left-most derivations will play an important role later.

**Lemma 1**  a) An Earley parser generates state

$$i: \quad {}_k X \rightarrow \lambda.\mu,$$

if and only if there is a partial derivation

$$S \stackrel{*}{\Rightarrow} x_{0\ldots k-1} X \nu \Rightarrow x_{0\ldots k-1} \lambda \mu \nu \stackrel{*}{\Rightarrow} x_{0\ldots k-1} x_{k\ldots i-1} \mu \nu$$

deriving a prefix $x_{0\ldots i-1}$ of the input.

b) There is a one-to-one mapping between partial derivations and Earley paths, such that each production $X \rightarrow \nu$ applied in a derivation corresponds to a predicted Earley state $X \rightarrow .\nu$.

(a) is the invariant underlying the correctness and completeness of Earley's algorithm; it can be proved by induction on the length of a derivation (Aho & Ullman 1972:Theorem 4.9). The slightly stronger form (b) follows from (a) and the way possible prediction steps are defined.

Since we have established that paths correspond to derivations, it is convenient to associate derivation probabilities directly with paths. The uniqueness condition (b) above, which is irrelevant to the correctness of a standard Earley parser, justifies (probabilistic) counting of paths in lieu of derivations.

**Definition 3** *The probability $P(\mathcal{P})$ of a path $\mathcal{P}$ is the product of the probabilities of all rules used in the predicted states occurring in $\mathcal{P}$.*

**Lemma 2**  a) For all paths $\mathcal{P}$ starting with a nonterminal $X$, $P(\mathcal{P})$ gives the probability of the (partial) derivation represented by $\mathcal{P}$. In particular, the string probability $P(X \stackrel{*}{\Rightarrow} x)$ is the sum of the probabilities of all paths starting with $X$ that are complete and constrained by $x$.

b) The sentence probability $P(S \stackrel{*}{\Rightarrow} x)$ is the sum of the probabilities of all complete paths starting with the initial state, constrained by $x$.

c) The prefix probability $P(S \stackrel{*}{\Rightarrow}_L x)$ is the sum of the probabilities of all paths $\mathcal{P}$ starting with the initial state, constrained by $x$, that end in a scanned state.

Note that when summing over all paths "starting with the initial state," summation is actually over all paths starting with $S$, by definition of the initial state $_0 \rightarrow .S$. (a) follows directly from our definitions of derivation probability, string probability, path probability and the one-to-one correspondence between paths and derivations established by Lemma 1. (b) follows from (a) by using $S$ as the start nonterminal. To obtain the prefix probability in (c), we need to sum the probabilities of all complete derivations that generate $x$ as a prefix. The constrained paths ending in scanned states represent exactly the beginnings of all such derivations. Since the grammar is assumed to be consistent and without useless nonterminals, all partial derivations can be completed with probability one. Hence the sum over the constrained incomplete paths is the sought-after sum over all complete derivations generating the prefix.

## 4.3  Forward and inner probabilities

Since string and prefix probabilities are the result of summing derivation probabilities, the goal is to compute these sums efficiently by taking advantage of the Earley control structure. This can be accomplished by attaching two probabilistic quantities to each Earley state, as follows. The terminology is derived from analogous or similar quantities commonly used in the literature on Hidden Markov Models (HMMs) (Rabiner & Juang 1986) and in Baker (1979).

**Definition 4** The following definitions are relative to an implied input string $x$.

a) The *forward probability* $\alpha_i(_kX \to \lambda.\mu)$ is the sum of the probabilities of all constrained paths of length $i$ that end in state $_kX \to \lambda.\mu$.

b) The *inner probability* $\gamma_i(_kX \to \lambda.\mu)$ is the sum of the probabilities of all paths of length $i - k$ that start in state $k : {}_kX \to .\lambda\mu$ and end in $i : {}_kX \to \lambda.\mu$, and generate the input symbols $x_k \ldots x_{i-1}$.

It helps to interpret these quantities in terms of an unconstrained Earley parser that operates as a generator emitting—rather than recognizing—strings. Instead of tracking all possible derivations, the generator traces along a single Earley path randomly determined by always choosing among prediction steps according to the associated rule probabilities. Notice that the scanning and completion steps are deterministic once the rules have been chosen.

Intuitively, the forward probability $\alpha_i(_kX \to \lambda.\mu)$ is the probability of an Earley generator producing the prefix of the input up to position $i - 1$ while passing through state $_kX \to \lambda.\mu$ at position $i$. However, due to left-recursion in productions the same state may appear several times on a path, and each occurrence is counted towards the total $\alpha_i$. Thus, $\alpha_i$ is really the *expected number* of occurrences of the given state in state set $i$. Having said that, we will refer to $\alpha$ simply as a probability, both for the sake of brevity, and to keep the analogy to the HMM terminology of which this is a generalization.[6] Note that for scanned states, $\alpha$ is always a probability, since by definition a scanned state can occur only once along a path.

The inner probabilities, on the other hand, represent the probability of generating a substring of the input from a given nonterminal, using a particular production. Inner probabilities are thus conditional on the presence of a given nonterminal $X$ with expansion starting at position $k$, unlike the forward probabilities, which include the generation history starting with the initial state. The inner probabilities as defined here correspond closely to the quantities of the same name in Baker (1979). The sum of $\gamma$ of all states with a given LHS $X$ is exactly Baker's inner probability for $X$.

The following is essentially a restatement of Lemma 2 in terms of forward and inner probabilities. It shows how to obtain the sentence and string probabilities we are interested in, provided that forward and inner probabilities can be computed effectively.

**Lemma 3** The following assumes an Earley chart constructed by the parser on an input string $x$ with $|x| = l$.

a) Provided that $S \overset{*}{\Rightarrow}_L x_{0 \ldots k-1} X \nu$ is a possible left-most derivation of the grammar (for some $\nu$), the probability that a nonterminal $X$ generates the substring $x_k \ldots x_{i-1}$ can be computed as the sum

$$P(X \overset{*}{\Rightarrow} x_{k \ldots i-1}) = \sum_{i : {}_kX \to \lambda.} \gamma_i(_kX \to \lambda.)$$

(sum of inner probabilities over all complete states with LHS $X$ and start index $k$).

b) In particular, the string probability $P(S \overset{*}{\Rightarrow} x)$ can be computed as[7]

$$\begin{aligned} P(S \overset{*}{\Rightarrow} x) &= \gamma_l(_0 \to S.) \\ &= \alpha_l(_0 \to S.) \end{aligned}$$

---

[6] The same technical complication was noticed by Wright (1990) in the computation of probabilistic LR parser tables. The relation to LR parsing will be discussed in Section 6.3. Incidentally, a similar interpretation of forward "probabilities" is required for HMMs with non-emitting states.

[7] The definitions of forward and inner probabilities coincide for the final state.

c) The prefix probability $P(S \stackrel{*}{\Rightarrow}_L x)$, with $|x| = l$, can be computed as

$$P(S \stackrel{*}{\Rightarrow}_L x) = \sum_{{}_k X \to \lambda x_{l-1}.\mu} \alpha_l({}_k X \to \lambda x_{l-1}.\mu)$$

(sum of forward probabilities over all scanned states).

The restriction in (a) that $X$ be preceded by a possible prefix is necessary since the Earley parser at position $i$ will only pursue derivations that are consistent with the input up to position $i$. This constitutes the main distinguishing feature of Earley parsing compared to the strict bottom-up computation used in the standard inside probability computation (Baker 1979). There, inside probabilities for all positions and nonterminals are computed, regardless of possible prefixes.

## 4.4 Computing forward and inner probabilities

Forward and inner probabilities not only subsume the prefix and string probabilities, they are also straightforward to compute during a run of Earley's algorithm. In fact, if it weren't for left-recursive and unit productions their computation would be trivial. For the purpose of exposition we will therefore ignore the technical complications introduced by these productions for a moment, and then return to them once the overall picture has become clear.

During a run of the parser both forward and inner probabilities will be attached to each state, and updated incrementally as new states are created through one of the three types of transitions. Both probabilities are set to unity for the initial state ${}_0 \to .S$. This is consistent with the interpretation that the initial state is derived from a dummy production $\to S$ for which no alternatives exist.

Parsing then proceeds as usual, with the probabilistic computations detailed below. The probabilities associated with new states will be computed as sums of various combinations of old probabilities. As new states are generated by prediction, scanning and completion, certain probabilities have to be *accumulated*, corresponding to the multiple paths leading to a state. That is, if the same state is generated multiple times, the previous probability associated with it has to be *incremented* by the new contribution just computed. States and probability contributions can be generated in any order, as long as the summation for one state is finished before its probability enters into the computation of some successor state. Appendix B.2 suggests a way to implement this incremental summation.

**Notation** A few intuitive abbreviations are used from here on to describe Earley transitions succinctly. (1) To avoid unwieldy $\sum$ notation we adopt the following convention. The expression $x \mathrel{+}= y$ means that $x$ is computed incrementally as a sum of various $y$ terms, which are computed in some order and accumulated to finally yield the value of $x$.[8] (2) Transitions are denoted by $\Longrightarrow$, with predecessor states on the left and successor states on the right. (3) The forward and inner probabilities of states are notated in brackets after each state, e.g.,

$$i: \quad {}_k X \to \lambda.Y\mu \quad [\alpha, \gamma]$$

is shorthand for $\alpha = \alpha_i({}_k X \to \lambda.Y\mu)$, $\gamma = \gamma_i({}_k X \to \lambda.Y\mu)$.

*Prediction (probabilistic)*

$$i: \quad {}_k X \to \lambda.Y\mu \quad [\alpha, \gamma] \quad \Longrightarrow \quad i: \quad {}_i Y \to .\nu \quad [\alpha', \gamma']$$

for all productions $Y \to \nu$. The new probabilities can be computed as

$$\begin{aligned} \alpha' &\mathrel{+}= \alpha \cdot P(Y \to \nu) \\ \gamma' &= P(Y \to \nu) \end{aligned}$$

---

[8] This notation suggests a simple implementation, being obviously borrowed from the programming language $C$.

Note that only the forward probability is accumulated; $\gamma$ is not used in this step.

*Rationale.* $\alpha'$ is the sum of all path probabilities leading up to $_kX \to \lambda.Y\mu$, times the probability of choosing production $Y \to \nu$. The value $\gamma'$ is just a special case of the definition.

*Scanning (probabilistic)*

$$i: \quad _kX \to \lambda.a\mu \quad [\alpha, \gamma] \quad \Longrightarrow \quad i+1: \quad _kX \to \lambda a.\mu \quad [\alpha', \gamma']$$

for all states with terminal $a$ matching input at position $i$. Then

$$\begin{aligned} \alpha' &= \alpha \\ \gamma' &= \gamma \end{aligned}$$

*Rationale.* Scanning does not involve any new choices since the terminal was already selected as part of the production during prediction.[9]

*Completion (probabilistic)*

$$\left. \begin{array}{l} i: \quad _jY \to \nu. \quad [\alpha'', \gamma''] \\ j: \quad _kX \to \lambda.Y\mu \quad [\alpha, \gamma] \end{array} \right\} \quad \Longrightarrow \quad i: \quad _kX \to \lambda Y.\mu \quad [\alpha', \gamma']$$

Then

$$\alpha' \mathrel{+}= \alpha \cdot \gamma'' \tag{1}$$
$$\gamma' \mathrel{+}= \gamma \cdot \gamma'' \tag{2}$$

Note that $\alpha''$ is not used.

*Rationale.* To update the old forward/inner probabilities $\alpha$ and $\gamma$ to $\alpha'$ and $\gamma'$, respectively, the probabilities of all paths expanding $Y \to \nu$ have to be factored in. These are exactly the paths summarized by the inner probability $\gamma''$.

## 4.5 Coping with recursion

The standard Earley algorithm, together with the probability computations described in the previous section would be sufficient if it weren't for the problem of recursion in the prediction and completion steps.

The non-probabilistic Earley algorithm can stop recursing as soon as all predictions/completions yield states already contained in the current state set. For the computation of probabilities, however, this would mean truncating the probabilities resulting from the repeated summing of contributions.

### 4.5.1 Prediction loops

As an example, consider the following simple left-recursive SCFG.

$$\begin{aligned} S &\to a \quad [p] \\ S &\to Sb \quad [q] \quad , \end{aligned}$$

---

[9] In different parsing scenarios the scanning step may well modify probabilities. For example, if the input symbols themselves have attached likelihoods these can be integrated by multiplying them onto $\alpha$ and $\gamma$ when a symbol is scanned. That way it is possible to perform efficient Earley parsing with integrated joint probability computation directly on weighted lattices describing ambiguous inputs.

where $q = 1 - p$. Non-probabilistically, the prediction loop at position 0 would stop after producing the states

$$
\begin{aligned}
_0 &\to .S \\
_0S &\to .a \\
_0S &\to .Sb
\end{aligned}
$$

This would leave the forward probabilities at

$$
\begin{aligned}
\alpha_0(_0S \to .a) &= p \\
\alpha_0(_0S \to .Sb) &= q \quad ,
\end{aligned}
$$

corresponding to just two out of an infinity of possible paths. The correct forward probabilities are obtained as a sum of infinitely many terms, accounting for all possible paths of length 1.

$$
\begin{aligned}
\alpha_0(_0S \to .a) &= p + qp + q^2p + \ldots = p(1-q)^{-1} = 1 \\
\alpha_0(_0S \to .Sb) &= q + q^2 + q^3 + \ldots = q(1-q)^{-1} = p^{-1}q
\end{aligned}
$$

In these sums each $p$ corresponds to a choice of the first production, each $q$ to a choice of the second production. If we didn't care about finite computation the resulting geometric series could be computed by letting the prediction loop (and hence the summation) continue indefinitely.

Fortunately, all repeated prediction steps, including those due to left-recursion in the productions, can be collapsed into a single, modified prediction step, and the corresponding sums computed in closed form. For this purpose we need a probabilistic version of the well-known parsing concept of a *left corner*, which is also at the heart of the prefix probability algorithm of Jelinek & Lafferty (1991).

**Definition 5** The following definitions are relative to a given SCFG $G$.

a) Two nonterminals $X$ and $Y$ are said to be in a *left-corner relation* $X \to_L Y$ iff there exists a production for $X$ that has a RHS starting with $Y$,

$$X \to Y\lambda \quad .$$

b) The *probabilistic left-corner relation*[10] $P_L = P_L(G)$ is the matrix of probabilities $P(X \to_L Y)$, defined as the total probability of choosing a production for $X$ that has $Y$ as a left corner:

$$P(X \to_L Y) = \sum_{X \to Y\lambda \in G} P(X \to Y\lambda) \quad .$$

c) The relation $X \stackrel{*}{\Rightarrow}_L Y$ is defined as the reflexive, transitive closure of $X \to_L Y$, i.e., $X \stackrel{*}{\Rightarrow}_L Y$ iff $X = Y$ or there is a nonterminal $Z$ such that $X \to_L Z$ and $Z \stackrel{*}{\Rightarrow}_L Y$.

d) The *probabilistic reflexive, transitive left-corner relation* $R_L = R_L(G)$ is a matrix of probability sums $R(X \stackrel{*}{\Rightarrow}_L Y)$. Each $R(X \stackrel{*}{\Rightarrow}_L Y)$ is defined as a series

$$
\begin{aligned}
R(X \stackrel{*}{\Rightarrow}_L Y) &= P(X = Y) \\
&+ P(X \to_L Y) \\
&+ \sum_{Z_1} P(X \to_L Z_1) P(Z_1 \to_P Y) \\
&+ \sum_{Z_1, Z_2} P(X \to_L Z_1) P(Z_1 \to_P Z_2) P(Z_2 \to_P Y) \\
&+ \ldots
\end{aligned}
$$

---

[10] If a *probabilistic relation* $R$ is replaced by its set-theoretic version $R'$, i.e., $(x, y) \in R'$ iff $R(x, y) \neq 0$, then the closure operations used here reduce to their traditional discrete counterparts; hence the choice of terminology.

Alternatively, $R_L$ is defined by the recurrence relation

$$R(X \overset{*}{\Rightarrow}_L Y) = \delta(X, Y) + \sum_Z P(X \to_L Z) R(Z \overset{*}{\Rightarrow}_L Y) \quad ,$$

where we use the delta function, defined as $\delta(X, Y) = 1$ if $X = Y$, and $\delta(X, Y) = 0$ if $X \neq Y$.

The recurrence for $R_L$ can be conveniently written in matrix notation

$$R_L = I + P_L R_L,$$

from which the closed-form solution is derived:

$$R_L = (I - P_L)^{-1}.$$

An existence proof for $R_L$ is given in Appendix A. Appendix B.3.1 shows how to speed up the computation of $R_L$ by inverting only a reduced version of the matrix $I - P_L$.

The significance of the matrix $R_L$ for the Earley algorithm is that its elements are the sums of the probabilities of the potentially infinitely many prediction paths leading from a state $_kX \to \lambda.Z\mu$ to a predicted state $_iY \to .\nu$, via any number of intermediate states.

$R_L$ can be computed once for each grammar, and used for table-lookup in the following, modified prediction step.

*Prediction (probabilistic, transitive)*

$$i: \quad _kX \to \lambda.Z\mu \quad [\alpha, \gamma] \quad \Longrightarrow \quad i: \quad _iY \to .\nu \quad [\alpha', \gamma']$$

for all productions $Y \to \nu$ such that $R(Z \overset{*}{\Rightarrow}_L Y)$ is non-zero. Then

$$\alpha' \mathrel{+}= \alpha \cdot R(Z \overset{*}{\Rightarrow}_L Y) P(Y \to \nu) \quad (3)$$
$$\gamma' = P(Y \to \nu) \quad (4)$$

The new $R(Z \overset{*}{\Rightarrow}_L Y)$ factor in the updated forward probability accounts for the sum of all path probabilities linking $Z$ to $Y$. For $Z = Y$ this covers the case of a single step of prediction; $R(Y \overset{*}{\Rightarrow}_L Y) \geq 1$ always, since $R_L$ is defined as a reflexive closure.

### 4.5.2 Completion loops

As in prediction, the completion step in the Earley algorithm may imply an infinite summation, and could lead to an infinite loop if computed naively. However, only *unit productions*[11] can give rise to cyclic completions.

The problem is best explained by studying an example. Consider the grammar

$$\begin{aligned} S &\to a \quad [p] \\ S &\to T \quad [q] \\ T &\to S \quad [1] \quad , \end{aligned}$$

where $q = 1 - p$. Presented with the input $a$ (the only string the grammar generates), after one cycle of prediction, the Earley chart contains the following states.

$$\begin{aligned} 0 &: {}_0 \to .S & \alpha &= 1, & \gamma &= 1 \\ 0 &: {}_0S \to .T & \alpha &= p^{-1}q, & \gamma &= q \\ 0 &: {}_0T \to .S & \alpha &= p^{-1}q, & \gamma &= 1 \\ 0 &: {}_0S \to .a & \alpha &= p^{-1}p = 1, & \gamma &= p. \end{aligned}$$

---

[11] Unit productions are also called "chain productions" or "single productions" in the literature.

The $p^{-1}$ factors are a result of the left-corner sum $1 + q + q^2 + \ldots = (1-q)^{-1}$.

After scanning $_0S \rightarrow .a$, completion without truncation would enter an infinite loop. First $_0T \rightarrow .S$ is completed, yielding a complete state $_0T \rightarrow S.$, which allows $_0S \rightarrow .T$ to be completed, leading to another complete state for $S$, etc. The non-probabilistic Earley parser can just stop here, but as in prediction, this would lead to truncated probabilities. The sum of probabilities that needs to be computed to arrive at the correct result contains infinitely many terms, one for each possible loop through the $T \rightarrow S$ production. Each such loop adds a factor of $q$ to the forward and inner probabilities. The summations for all completed states turn out as

$$
\begin{array}{rlll}
1 : {}_0S & \rightarrow & x. & \alpha = 1, \quad \gamma = p \\
1 : {}_0T & \rightarrow & S. & \alpha = p^{-1}q(p + pq + pq^2 + \ldots) = p^{-1}q, \quad \gamma = p + pq + pq^2 + \ldots = 1 \\
1 : {}_0 & \rightarrow & S. & \alpha = p + pq + pq^2 + \ldots = 1, \quad \gamma = p + pq + pq^2 + \ldots = 1 \\
1 : {}_0S & \rightarrow & T. & \alpha = p^{-1}q(p + pq + pq^2 + \ldots) = p^{-1}q, \quad \gamma = q(p + pq + pq^2 + \ldots) = q
\end{array}
$$

The approach taken here to compute exact probabilities in cyclic completions is mostly analogous to that for left-recursive predictions. The main difference is that unit productions, rather than left-corners, form the underlying transitive relation. Before proceeding we can convince ourselves that this is indeed the only case we have to worry about.

**Lemma 4** Let

$$_{k_1}X_1 \rightarrow \lambda_1 X_2. \Longrightarrow {}_{k_2}X_2 \rightarrow \lambda_2 X_3. \Longrightarrow \ldots \Longrightarrow {}_{k_c}X_c \rightarrow \lambda_c X_{c+1}.$$

be a completion cycle, i.e., $k_1 = k_c$, $X_1 = X_c$, $\lambda_1 = \lambda_c$, $X_2 = X_{c+1}$. Then it must be the case that $\lambda_1 = \lambda_2 = \ldots = \lambda_c = \epsilon$, i.e., all productions involved are unit productions $X_1 \rightarrow X_2, \ldots, X_c \rightarrow X_{c+1}$.

*Proof.* For all completion chains it is true that the start indices of the states are monotonically increasing, $k_1 \geq k_2 \geq \ldots$ (a state can only complete an expansion that started at the same or a previous position). From $k_1 = k_c$ it follows that $k_1 = k_2 = \ldots = k_c$. Because the current position (dot) also refers to the same input index in all states, all nonterminals $X_1, X_2, \ldots, X_c$ have been expanded into the same substring of the input between $k_1$ and the current position. By assumption the grammar contains no nonterminals that generate $\epsilon$,[12] therefore we must have $\lambda_1 = \lambda_2 = \ldots = \lambda_c = \epsilon$, q.e.d.

We now formally define the relation between nonterminals mediated by unit productions, analogous to the left-corner relation.

**Definition 6** The following definitions are relative to a given SCFG $G$.

a) Two nonterminals $X$ and $Y$ are said to be in a *unit-production relation* $X \rightarrow Y$ iff there exists a production for $X$ that has $Y$ as its RHS.

b) The *probabilistic unit-production relation* $P_U = P_U(G)$ is the matrix of probabilities $P(X \rightarrow Y)$.

c) The relation $X \overset{*}{\Rightarrow} Y$ is defined as the reflexive, transitive closure of $X \rightarrow Y$, i.e., $X \overset{*}{\Rightarrow} Y$ iff $X = Y$ or there is a nonterminal $Z$ such that $X \rightarrow Z$ and $Z \overset{*}{\Rightarrow} Y$.

---

[12]Even with null productions, these would not be used for Earley transitions, see Section 4.7.

d) The *probabilistic reflexive, transitive unit-production relation* $R_U = R_U(G)$ is the matrix of probability sums $R(X \stackrel{*}{\Rightarrow} Y)$. Each $R(X \stackrel{*}{\Rightarrow} Y)$ is defined as a series

$$\begin{aligned}
R(X \stackrel{*}{\Rightarrow} Y) &= P(X = Y) \\
&\quad + P(X \to Y) \\
&\quad + \sum_{Z_1} P(X \to Z_1) P(Z_1 \to Y) \\
&\quad + \sum_{Z_1, Z_2} P(X \to Z_1) P(Z_1 \to Z_2) P(Z_2 \to Y) \\
&\quad + \ldots \\
&= \delta(X, Y) + \sum_Z P(X \to Z) R(Z \stackrel{*}{\Rightarrow} Y) \quad .
\end{aligned}$$

As before, a matrix inversion can compute the relation $R_U$ in closed form:

$$R_U = (I - P_U)^{-1} .$$

The existence of $R_U$ is shown in Appendix A.

The modified completion loop in the probabilistic Earley parser can now use the $R_U$ matrix to collapse all unit completions into a single step. Note that we still have to do iterative completion on non-unit productions.

*Completion (probabilistic, transitive)*

$$\left. \begin{array}{ll} i: & {}_j Y \to \nu. \quad [\alpha'', \gamma''] \\ j: & {}_k X \to \lambda.Z\mu \quad [\alpha, \gamma] \end{array} \right\} \quad \Longrightarrow \quad i: \quad {}_k X \to \lambda Z.\mu \quad [\alpha', \gamma']$$

for all $Y, Z$ such that $R(Z \stackrel{*}{\Rightarrow} Y)$ is non-zero, and $Y \to \nu$ is not a unit production ($|\nu| > 1$ or $\nu \in \Sigma$). Then

$$\begin{aligned}
\alpha' &\mathrel{+}= \alpha \cdot \gamma'' R(Z \stackrel{*}{\Rightarrow} Y) \\
\gamma' &\mathrel{+}= \gamma \cdot \gamma'' R(Z \stackrel{*}{\Rightarrow} Y)
\end{aligned}$$

### 4.6 An example

Consider the grammar

$$\begin{aligned}
S &\to a \quad [p] \\
S &\to SS \quad [q]
\end{aligned}$$

where $q = 1 - p$. This highly ambiguous grammar generates strings of any number of $a$'s, using all possible binary parse trees over the given number of terminals. The states involved in parsing the string $aaa$ are listed in Table 2, along with their forward and inner probabilities. The example illustrates how the parser deals with left-recursion and the merging of alternative sub-parses during completion.

Since the grammar has only a single nonterminal, the left-corner matrix $P_L$ has rank 1:

$$P_L = [q] \quad .$$

Its transitive closure is

$$R_L = (I - P_L)^{-1} = [p]^{-1} = [p^{-1}] \quad .$$

(a)

$$S \rightarrow a \quad [p]$$
$$S \rightarrow SS \quad [q]$$

(b)

|  | $\alpha$ | $\gamma$ |
|---|---|---|
| State set 0 | | |
| $_0 \rightarrow .S$ | 1 | 1 |
| *predicted* | | |
| $_0S \rightarrow .a$ | $1 \cdot p^{-1}p = 1$ | $p$ |
| $_0S \rightarrow .SS$ | $1 \cdot p^{-1}q = p^{-1}q$ | $q$ |
| State set 1 | | |
| *scanned* | | |
| $_0S \rightarrow a.$ | $p^{-1}p = 1$ | $p$ |
| *completed* | | |
| $_0S \rightarrow S.S$ | $p^{-1}q \cdot p = q$ | $q \cdot p = pq$ |
| *predicted* | | |
| $_1S \rightarrow .a$ | $q \cdot p^{-1}p = q$ | $p$ |
| $_1S \rightarrow .SS$ | $q \cdot p^{-1}q = p^{-1}q^2$ | $q$ |
| State set 2 | | |
| *scanned* | | |
| $_1S \rightarrow a.$ | $q$ | $p$ |
| *completed* | | |
| $_1S \rightarrow S.S$ | $p^{-1}q^2 \cdot p = q^2$ | $q \cdot q = pq$ |
| $_0S \rightarrow SS.$ | $q \cdot p = pq$ | $pq \cdot p = p^2q$ |
| $_0S \rightarrow S.S$ | $p^{-1}q \cdot p^2q = pq^2$ | $q \cdot p^2q = p^2q^2$ |
| $_0 \rightarrow S.$ | $1 \cdot p^2q = p^2q$ | $1 \cdot p^2q = p^2q$ |
| *predicted* | | |
| $_2S \rightarrow .a$ | $(q^2 + pq^2) \cdot p^{-1}p = (1+p)q^2$ | $p$ |
| $_2S \rightarrow .SS$ | $(q^2 + pq^2) \cdot p^{-1}q = (1+p^{-1})q^3$ | $q$ |
| State set 3 | | |
| *scanned* | | |
| $_2S \rightarrow a.$ | $(1+p)q^2$ | $p$ |
| *completed* | | |
| $_2S \rightarrow S.S$ | $(1+p^{-1})q^3 \cdot p = (1+p)q^3$ | $q \cdot p = pq$ |
| $_1S \rightarrow SS.$ | $q^2 \cdot p = pq^2$ | $pq \cdot p = p^2q$ |
| $_1S \rightarrow S.S$ | $p^{-1}q^2 \cdot p^2q = pq^3$ | $q \cdot p^2q = p^2q^2$ |
| $_0S \rightarrow SS.$ | $pq^2 \cdot p + q \cdot p^2q = 2p^2q^2$ | $p^2q^2 \cdot p + pq \cdot p^2q = 2p^3q^2$ |
| $_0S \rightarrow S.S$ | $p^{-1}q \cdot 2p^3q^2 = 2p^2q^3$ | $q \cdot 2p^3q^2 = 2p^3q^3$ |
| $_0 \rightarrow S.$ | $1 \cdot 2p^3q^2 = 2p^3q^2$ | $1 \cdot 2p^3q^2 = 2p^3q^2$ |

Table 2: Earley chart as constructed during the parse of *aaa* with the grammar in (a). The two columns to the right in (b) list the forward and inner probabilities, respectively, for each state. In both $\alpha$ and $\gamma$ columns, the $\cdot$ separates old factors from new ones (as per equations 1, 2 and 3). Addition indicates multiple derivations of the same state.

Consequently, the example trace shows the factor $p^{-1}$ being introduced into the forward probability terms in the prediction steps.

The sample string can be parsed as either $(a(aa))$ or $((aa)a)$, each parse having a probability of $p^3q^2$. The total string probability is thus $2p^3q^2$, the computed $\alpha$ and $\gamma$ values for the final state. The $\alpha$ values for the scanned states in sets 1, 2 and 3 are the prefix probabilities for $a$, $aa$, and $aaa$, respectively: $P(S \stackrel{*}{\Rightarrow}_L a) = 1$, $P(S \stackrel{*}{\Rightarrow}_L aa) = q$, $P(S \stackrel{*}{\Rightarrow}_L aaa) = (1+p)q^2$.

## 4.7 Null productions

Null productions $X \rightarrow \epsilon$ introduce some complications into the relatively straightforward parser operation described so far, some of which are due specifically to the probabilistic aspects of parsing. This section summarizes the necessary modifications to process null productions correctly, using the previous description as a baseline. Our treatment of null productions follows the (non-probabilistic) formulation of Graham *et al.* (1980), rather than the original one in Earley (1970).

### 4.7.1 Computing $\epsilon$-expansion probabilities

The main problem with null productions is that they allow multiple prediction-completion cycles inbetween scanning steps (since null productions do not have to be matched against one or more input symbols). Our strategy will be to collapse all predictions and completions due to chains of null productions into the regular prediction and completion steps, not unlike the way recursive predictions/completions were handled in Section 4.5.

A prerequisite for this approach is to precompute, for all nonterminals $X$, the probability that $X$ expands to the empty string. Note that this is another recursive problem, since $X$ itself may not have a null production, but expand to some nonterminal $Y$ that does.

Computation of $P(X \stackrel{*}{\Rightarrow} \epsilon)$ for all $X$ can be cast as a system of non-linear equations, as follows. For each $X$, let $e_X$ be an abbreviation for $P(X \stackrel{*}{\Rightarrow} \epsilon)$. For example, let $X$ have productions

$$
\begin{aligned}
X &\rightarrow \epsilon & [p_1] \\
&\rightarrow Y_1 Y_2 & [p_2] \\
&\rightarrow Y_3 Y_4 Y_5 & [p_3] \\
&\vdots
\end{aligned}
$$

The semantics of context-free rules imply that $X$ can only expand to $\epsilon$ if *all* the RHS nonterminals in one of $X$'s productions expand to $\epsilon$. Translating to probabilities, we obtain the equation

$$e_X = p_1 + p_2 e_{Y_1} e_{Y_2} + p_3 e_{Y_3} e_{Y_4} e_{Y_5} + \ldots$$

In other words, each production contributes a term in which the rule probability is multiplied by the product of the $e$ variables corresponding to the RHS nonterminals, unless the RHS contains a terminal (in which case the production contributes nothing to $e_X$ because it cannot possibly lead to $\epsilon$).

The resulting non-linear system can be solved by iterative approximation. Each variable $e_X$ is initialized to $P(X \rightarrow \epsilon)$, and then repeatedly updated by substituting in the equation right-hand sides, until the desired level of accuracy is attained. Convergence is guaranteed since the $e_X$ values are monotonically increasing and bounded above by the true values $P(X \stackrel{*}{\Rightarrow} \epsilon) \leq 1$. For grammars without cyclic dependencies among $\epsilon$-producing nonterminals this procedure degenerates to simple backward substitution. Obviously the system has to be solved only once for each grammar.

The probability $e_X$ can be seen as the precomputed inner probability of an expansion of $X$ to the empty string, i.e., it sums the probabilities of all Earley paths that derive $\epsilon$ from $X$. This is the justification for the way these probabilities can be used in modified prediction and completion steps, described next.

### 4.7.2 Prediction with null productions

Prediction is mediated by the left-corner relation. For each $X$ occurring to the right of a dot, we generate states for all $Y$ that are reachable from $X$ by way of the $X \rightarrow_L Y$ relation. This reachability criterion has to be extended in the presence of null productions. Specifically, if $X$ has a production $X \rightarrow Y_1 \ldots Y_{i-1} Y_i \lambda$ then $Y_i$ is a left corner of $X$ iff $Y_1, \ldots, Y_{i-1}$ all have a non-zero probability of expanding to $\epsilon$. The contribution of such a production to the left-corner probability $P(X \rightarrow_L Y_i)$ is

$$P(X \rightarrow Y_1 \ldots Y_{i-1} Y_i \lambda) \prod_{k=1}^{i-1} e_{Y_k}$$

The old prediction procedure can now be modified in two steps. First, replace the old $P_L$ relation by the one that takes into account null productions, as sketched above. From the resulting $P_L$ compute the reflexive transitive closure $R_L$, and use it to generate predictions as before.

Second, when predicting a left corner $Y$ with a production $Y \rightarrow Y_1 \ldots Y_{i-1} Y_i \lambda$, add states for all dot positions up to the first RHS nonterminal that cannot expand to $\epsilon$, say from $X \rightarrow .Y_1 \ldots Y_{i-1} Y_i \lambda$ through $X \rightarrow Y_1 \ldots Y_{i-1}.Y_i \lambda$. We will call this procedure "spontaneous dot shifting." It accounts precisely for those derivations that expand the RHS prefix $Y_1 \ldots Y_{i-1}$ without consuming any of the input symbols.

The forward and inner probabilities of the states thus created are those of the first state $X \rightarrow .Y_1 \ldots Y_{i-1} Y_i \lambda$, multiplied by factors that account for the implied $\epsilon$-expansions. This factor is just the product $\prod_{k=1}^{j} e_{Y_k}$, where $j$ is the dot position.

### 4.7.3 Completion with null productions

Modification of the completion step follows a similar pattern. First, the unit-production relation has to be extended to allow for unit-production chains due to null productions. A rule $X \rightarrow Y_1 \ldots Y_{i-1} Y_i Y_{i+1} \ldots Y_j$ can effectively act as a unit production that links $X$ and $Y_i$ if all other nonterminals on the RHS can expand to $\epsilon$. Its contribution to the unit production relation $P(X \rightarrow Y_i)$ will then be

$$P(X \rightarrow Y_1 \ldots Y_{i-1} Y_i Y_{i+1} \ldots Y_j) \prod_{k \neq i} e_{Y_k}$$

From the resulting revised $P_U$ matrix we compute the closure $R_U$ as usual.

The second modification is another instance of spontaneous dot shifting. When completing a state $X \rightarrow \lambda.Y\mu$ and moving the dot to get $X \rightarrow \lambda Y.\mu$, additional states have to be added, obtained by moving the dot further over any nonterminals in $\mu$ that have non-zero $\epsilon$-expansion probability. As in prediction, forward and inner probabilities are multiplied by the corresponding $\epsilon$-expansion probabilities.

### 4.7.4 Eliminating null productions

Given these added complications one might consider simply eliminating all $\epsilon$-productions in a pre-processing step. This is mostly straightforward and analogous to the corresponding procedure for non-probabilistic CFGs (Aho & Ullman 1972:Algorithm 2.10). The main difference is the updating of rule probabilities, for which the $\epsilon$-expansion probabilities are again needed.

1. Delete all null productions, except on the start symbol (in case the grammar as a whole produces $\epsilon$ with non-zero probability). Scale the remaining production probabilities to sum to unity.

2. For each original rule $X \rightarrow \lambda Y \mu$ that contains a nonterminal $Y$ such that $Y \stackrel{*}{\Rightarrow} \epsilon$:

    (a) Create a variant rule $X \rightarrow \lambda \mu$

(b) Set the rule probability of the new rule to $e_Y P(X \to \lambda Y \mu)$. If the rule $X \to \lambda \mu$ already exists, sum the probabilities.

(c) Decrement the old rule probability by the same amount.

Iterate these steps for all RHS occurrences of a null-able nonterminal.

The crucial step in this procedure is the addition of variants of the original productions that simulate the null productions by deleting the corresponding nonterminals from the RHS. The spontaneous dot shifting described in the previous sections effectively performs the same operation on the fly as the rules are used in prediction and completion.

## 4.8 Complexity issues

The probabilistic extension of Earley's parser preserves the original control structure in most aspects, the major exception being the collapsing of cyclic predictions and unit completions, which can only make these steps more efficient. We can therefore apply the complexity analysis from Earley (1970) essentially unchanged. Below we repeat the highlights, together with proof outlines. We also analyze the dependence on the size of the grammar, and compare the result with the other known algorithms for SCFGs.

The key factor in upper-bounding both time and space complexity in Earley's algorithm is the maximal number of states created in each state set, for each input position. An Earley state combines a production, a dot position, and a start index. Productions and dot positions combine to give a number that equals the sum of the lengths of all productions, which is roughly the total "size" of the grammar. For fully parameterized CNF grammars the number of dotted rules is $O(n^3)$, where $n$ is the number of nonterminals. (A fully parameterized CNF grammar is one in which each triple of nonterminals $X, Y, Z$ forms a production $X \to YZ$ of non-zero probability.) In both cases the start index contributes a factor of at most $l$, the length of the input.

### 4.8.1 Scaling with input length

To determine the complexity in terms of $l$ we note that during prediction and scanning, each state is processed exactly once, performing operations that depend only on the size of the grammar, not $l$; both therefore take $O(l)$. During completion, the worst case is obtained if each of the $O(l)$ states is the result of completing predecessors (with dot positions further left) from all possible previous positions. The total time taken here is thus $O(l^2)$. The completion step thus dominates the computation time, and gives $O(l^3)$ total run time over the entire string.

Earley (1970) identifies several important classes of context-free grammars on which the algorithm runs faster without special modifications. Grammars with no or bounded ambiguity result in completions that have to combine at most a fixed number of previous states (combining completions correspond to coalescing multiple parses for substrings). Such a completion step therefore takes time $O(l)$, to give a total of $O(l^2)$. CFGs that can be processed deterministically, i.e., where the correct choice of rule can be determined using only the history of the parser and a bounded lookahead into the input, such as LR($k$) grammars, result in an Earley chart that contains only a fixed maximum number of states per position. (To realize the benefit of deterministic parsing one generally needs the lookahead feature of Earley's original version, not discussed here.) Prediction, scanning and completion all take constant time (in terms of $l$) in this case, so the overall time complexity is linear.

If our parser is to be used for on-line computation of prefix probabilities it is critical to know the incremental time complexity for the processing of the next input symbol (or word). From the analysis above we get a worst case incremental time of $O(l^2)$, $O(l)$ for bounded ambiguity grammars, and constant time for deterministic grammars. Since $l$ in this case actually refers to the length of the prefix, incremental processing is generally slowed down as more of the input is incorporated in the chart.

The space complexity in terms of $l$ is $O(l^2)$ since $l$ state sets of $O(l)$ elements each have to be created.

All in all, we get the same time $O(l^3)$, space $O(l^2)$ bounds as in the Inside/Outside (Baker 1979) and LRI (Jelinek & Lafferty 1991) algorithms, with the advantage of better results on known grammar classes.

### 4.8.2 Scaling with grammar size

We will not try to give a precise characterization in the case of sparse grammars (Appendix B.3 gives some hints on how to implement the algorithm efficiently for such grammars). However, for fully parameterized grammars in CNF we can verify the scaling of our algorithm in terms of the number of nonterminals $n$, and compare it to the I/O and LRI algorithms, which both run in time $O(n^3)$.

As already mentioned, the number of states per position is $O(ln^3)$ for a CNF grammar. During prediction, summation of forward probabilities (equation 3) can be implemented efficiently as follows. We first compute the sum of all $\alpha$'s referring to a given nonterminal $X$ right of the dot, for all $X$. This can be done in a single pass over the current state set, i.e., in time $O(ln^3)$. The result is a vector of $\alpha$-sums, indexed by nonterminals. Multiplying this vector with the matrix $R_L$ we get another vector (in time $O(n^2)$). The $\alpha'$ from equation (3) are obtained by multiplying the rule probability $P(Y \rightarrow \nu)$ with the $Y$ element in that vector (total time for this step $O(n)$).

Scanning involves shifting the dot in the $O(ln)$ states that represent terminal productions. During completion we again have to update probabilities for $O(ln^3)$ states, each of which is the result of summing over $O(l)$ predecessors. (Note that there can be no cyclic completions with CNF grammars.) To implement summations (1) and (2) efficiently we first sum the inner probabilities $\gamma''$ from all states that refer to the same LHS nonterminal in a single $O(n^3)$ pass.

Finally, the matrix inversion to compute the left-corner and unit-production relation matrices is also accomplished in $O(n^3)$ time. However, that cost can be amortized over all subsequent uses of the parser.

The space requirements of all algorithms discussed here are proportional to the number of parameters, i.e., $O(n^3)$.

Overall, we get the same $O(n^3)$ dependence on the number of nonterminals as for the I/O and LRI algorithm.

## 4.9 Summary

To summarize, the modified, probabilistic Earley algorithm works by executing the following steps for each input position.

- Apply a single prediction step to all incomplete states in the current state set. All transitive predictions are subsumed by consulting the left-corner matrix $R_L$.

  Forward probabilities are computed by multiplying old $\alpha$'s with rule probabilities. Inner probabilities are initialized to their respective rule probabilities.

- A single scanning step applied to all states with terminals to the right of the dot yield the initial elements for the next state set. If the next set remains empty (no scanned states) the parse is aborted.

  Forward and inner probabilities remain unchanged by scanning.

  The sum of all forward probabilities of successfully scanned states gives the current prefix probability.

- Apply iterative completion (highest start index first, breadth-first) to all states, except those corresponding to unit productions. Unit production cycles are subsumed by consulting the matrix $R_U$.

  Forward and inner probabilities are updated by multiplying old forward and inner probabilities with the inner probabilities of completed expansions.

  The probabilities that nonterminals $X$ generate particular substrings of the input can be computed as sums of inner probabilities $\gamma(_kX \to \lambda.)$

After processing the entire string in this way, the sentence probability can be read off of either the $\alpha$ or $\gamma$ of the final state.

# 5 Extensions

This section discusses extensions to the Earley algorithm that go beyond simple parsing and the computation of prefix and string probabilities. These extension are all quite straightforward and well-supported by the original Earley chart structure, which leads us to view them as part of a single, unified algorithm for solving the tasks mentioned in the introduction.

## 5.1 Viterbi parses

**Definition 7** A *Viterbi parse* for a string $x$, in a grammar $G$, is a left-most derivation that assigns maximal probability to $x$, among all possible derivations for $x$.

Both the definition of Viterbi parse, and its computation are straightforward generalizations of the corresponding notion for Hidden Markov Models (Rabiner & Juang 1986), where one computes the Viterbi *path* (state sequence) through an HMM. Precisely the same approach can be used in the Earley parser, using the fact that each derivation corresponds to a path.

The standard computational technique for Viterbi parses is applicable here. Wherever the original parsing procedure sums probabilities that correspond to alternative derivations of a grammatical entity, the summation is replaced by a maximization. Thus, during the forward pass each state must keep track of the maximal path probability leading to it, as well as the predecessor states associated with that maximum probability path. Once the final state is reached, the maximum probability parse can be recovered by tracing back the path of "best" predecessor states.

The following modifications to the probabilistic Earley parser implement the forward phase of the Viterbi computation.

- Each state computes an additional probability, its *Viterbi probability* $v$.

- Viterbi probabilities are propagated in the same way as inner probabilities, except that during completion the summation is replaced by maximization: $v_i(_kX \to \lambda Y.\mu)$ is the maximum of all products $v_i(_jY \to \nu.)v_j(_kX \to \lambda.Y\mu)$ that contribute to the completed state $_kX \to \lambda Y.\mu$. The same-position predecessor $_jY \to \nu.$ associated with the maximum is recorded as the Viterbi path predecessor of $_kX \to \lambda Y.\mu$ (the other predecessor state $_kX \to \lambda.Y\mu$ can be inferred).

- The completion step uses the original recursion without collapsing of unit production loops. Loops are simply avoided, since they can only lower a path's probability. Collapsing of unit-production completions has to be avoided to maintain a continuous chain of predecessors for later backtracing and parse construction.

- The prediction step does not need to be modified for the Viterbi computation.

Once the final state is reached, a recursive procedure can recover the parse tree associated with the Viterbi parse. This procedure takes an Earley state $i : {}_kX \rightarrow \lambda.\mu$ as input and produces the Viterbi parse for the substring between $k$ and $i$ as output. (If the input state is not complete ($\mu \neq \epsilon$), the result will be a partial parse tree with children missing from the root node.)

**Viterbi-parse**($i : {}_kX \rightarrow \lambda.\mu$):

1. If $\lambda = \epsilon$, return a parse tree with root labeled $X$ and no children.

2. Otherwise, if $\lambda$ ends in a terminal $a$, let $\lambda'a = \lambda$, and call this procedure recursively to obtain the parse tree
$$T = \text{Viterbi-parse}(i - 1 : {}_kX \rightarrow \lambda'.a\mu)$$
Adjoin a leaf node labeled $a$ as the right-most child to the root of $T$ and return $T$.

3. Otherwise, if $\lambda$ ends in a nonterminal $Y$, let $\lambda'Y = \lambda$. Find the Viterbi predecessor state ${}_jY \rightarrow \nu.$ for the current state. Call this procedure recursively to compute
$$T = \text{Viterbi-parse}(j : {}_kX \rightarrow \lambda'.Y\mu)$$
as well as
$$T' = \text{Viterbi-parse}(i : {}_jY \rightarrow \nu.)$$
Adjoin $T'$ to $T$ as the right-most child at the root, and return $T$.

## 5.2 Rule probability estimation

The rule probabilities in a SCFG can be iteratively estimated using the EM (Expectation-Maximization) algorithm (Dempster *et al.* 1977). Given a sample corpus $D$, the estimation procedure finds a set of parameters that represent a local maximum of the grammar likelihood function $P(D|G)$, which is given by the product of the string probabilities

$$P(D|G) = \prod_{x \in D} P(S \stackrel{*}{\Rightarrow} x) \quad ,$$

i.e., the samples are assumed to be distributed identically and independently.

The two steps of this algorithm can be briefly characterized as follows.

**E-step:** Compute expectations for how often each grammar rule is used, given the corpus $D$ and the current grammar parameters (rule probabilities).

**M-step:** Reset the parameters so as to maximize the likelihood relative to the expected rule counts found in the E-step.

This procedure is iterated until the parameter values (as well as the likelihood) converge. It can be shown that each round in the algorithm produces a likelihood that is a least a high as the previous one; the EM algorithm is therefore guaranteed to find at least a local maximum of the likelihood function.

EM is a generalization of the well-known Baum-Welch algorithm for HMM estimation (Baum *et al.* 1970); the original formulation for the case of SCFGs is due to Baker (1979). For SCFGs, the E-step involves computing the expected number of times each production is applied in generating the training corpus. After that, the M-step consists of a simple normalization of these counts to yield the new production probabilities.

In this section we examine the computation of production count expectations required for the E-step. The crucial notion introduced by Baker (1979) for this purpose is the "outer probability" of a nonterminal, or the joint probability that the nonterminal is generated with a given prefix and suffix of terminals. Essentially the same method can be used in the Earley framework, after extending the definition of outer probabilities to apply to arbitrary Earley states.

**Definition 8** Given a string $x$, $|x| = l$, the *outer probability* $\beta_i({}_kX \to \lambda.\mu)$ of an Earley state is the sum of the probabilities of all paths that

- start with the initial state,
- generate the prefix $x_0 \ldots x_{k-1}$,
- pass through ${}_kX \to .\nu\mu$, for some $\nu$,
- generate the suffix $x_i \ldots x_{l-1}$ starting with state ${}_kX \to \nu.\mu$,
- end in the final state.

Outer probabilities complement inner probabilities in that they refer precisely to those parts of complete paths generating $x$ not covered by the corresponding inner probability $\gamma_i({}_kX \to \lambda.\mu)$. Therefore, the choice of the production $X \to \lambda\mu$ is *not* part of the outer probability associated with a state ${}_kX \to \lambda.\mu$. In fact, the definition makes no reference to the first part $\lambda$ of the RHS: all states sharing the same $k$, $X$ and $\mu$ will have identical outer probabilities.

Intuitively, $\beta_i({}_kX \to \lambda.\mu)$ is the probability that an Earley parser operating as a string generator yields the prefix $x_{0\ldots k-1}$ and the suffix $x_{i\ldots l-1}$, while passing through state ${}_kX \to \lambda.\mu$ at position $i$ (which is independent of $\lambda$). As was the case for forward probabilities, $\beta$ is actually an expectation of the number of such states in the path, as unit production cycles can result in multiple occurrences for a single state. Again, we gloss over this technicality in our terminology. The name is motivated by the fact that $\beta$ reduces to the "outer probability" of $X$ as defined in Baker (1979) if the dot is in final position.

### 5.2.1 Computing expected production counts

Before going into the details of computing outer probabilities we describe their use in obtaining the expected rule counts needed for the E-step in grammar estimation.

Let $c(X \to \lambda | x)$ denote the expected number of uses of production $X \to \lambda$ in the derivation of string $x$. Alternatively, $c(X \to \lambda | x)$ is the expected number of times that $X \to \lambda$ is used for prediction in a complete Earley path generating $x$. Let $c(X \to \lambda | \mathcal{P})$ be the number of occurrences of predicted states based on production $X \to \lambda$ along a path $\mathcal{P}$.

$$
\begin{aligned}
c(X \to \lambda | x) &= \sum_{\mathcal{P} \text{ derives } x} P(\mathcal{P} | S \stackrel{*}{\Rightarrow} x) c(X \to \lambda | \mathcal{P}) \\
&= \frac{1}{P(S \stackrel{*}{\Rightarrow} x)} \sum_{\mathcal{P} \text{ derives } x} P(\mathcal{P}, S \stackrel{*}{\Rightarrow} x) c(X \to \lambda | \mathcal{P}) \\
&= \frac{1}{P(S \stackrel{*}{\Rightarrow} x)} \sum_{i : {}_iX \to .\lambda} P(S \stackrel{*}{\Rightarrow} x_{0 \ldots i-1} X \nu \stackrel{*}{\Rightarrow} x) \quad .
\end{aligned}
$$

The last summation is over all predicted states based on production $X \to \lambda$. The quantity $P(S \stackrel{*}{\Rightarrow} x_{0 \ldots i-1} X \nu \stackrel{*}{\Rightarrow} x)$ is the sum of the probabilities of all paths passing through $i : {}_iX \to .\lambda$. Inner and outer probabilities have been defined such that this quantity is obtained precisely as the product of the corresponding of $\gamma_i$ and $\beta_i$. Thus, the expected usage count for a rule can be computed as

$$
c(X \to \lambda | x) = \frac{1}{P(S \stackrel{*}{\Rightarrow} x)} \sum_{i : {}_iX \to .\lambda} \beta_i({}_iX \to .\lambda) \gamma_i({}_iX \to .\lambda) \quad .
$$

The sum can be computed after completing both forward and backward passes (or during the backward pass itself) by scanning the chart for predicted states.

### 5.2.2 Computing outer probabilities

The outer probabilities are computed by tracing the complete paths from the final state to the start state, in a single backward pass over the Earley chart. Only completion and scanning steps need to be traced back. Reverse scanning leaves outer probabilities unchanged, so the only operation of concern is reverse completion.

We describe reverse transitions using the same notation as for their forward counterparts, annotating each state with its outer and inner probabilities.

*Reverse completion*

$$i: \quad {}_kX \to \lambda Y.\mu \quad [\beta, \gamma] \quad \Longrightarrow \quad \begin{cases} i: \quad {}_jY \to \nu. \quad [\beta'', \gamma''] \\ j: \quad {}_kX \to \lambda.Y\mu \quad [\beta', \gamma'] \end{cases}$$

for all pairs of states ${}_jY \to \nu.$ and ${}_kX \to \lambda.Y\mu$ in the chart. Then

$$\begin{aligned} \beta' &\mathrel{+}= \gamma'' \cdot \beta \\ \beta'' &\mathrel{+}= \gamma' \cdot \beta \end{aligned}$$

The inner probability $\gamma$ is not used.

*Rationale.* Relative to $\beta'$, $\beta$ is missing the probability of expanding $Y$, which is filled in from $\gamma''$. The probability of the surrounding of $Y$ is the probability of the surrounding of $X$, plus the choice of the rule of production for $X$ and the expansion of the partial LHS $\lambda$, which are together given by $\gamma'$.

Note that the computation makes use of the inner probabilities computed in the forward pass. The particular way in which $\gamma$ and $\beta$ were defined turns out to be convenient here, as no reference to the production probabilities themselves needs to be made in the computation.

As in the forward pass, simple reverse completion would not terminate in the presence of cyclic unit productions. A version that collapses all such chains of productions is given below.

*Reverse completion (transitive)*

$$i: \quad {}_kX \to \lambda Z.\mu \quad [\beta, \gamma] \quad \Longrightarrow \quad \begin{cases} i: \quad {}_jY \to \nu. \quad [\beta'', \gamma'] \\ j: \quad {}_kX \to \lambda.Z\mu \quad [\beta', \gamma'] \end{cases}$$

for all pairs of states ${}_jY \to \nu.$ and ${}_kX \to \lambda.Z\mu$ in the chart, such that the unit-production relation $R(Z \overset{*}{\Rightarrow} Y)$ is non-zero. Then

$$\begin{aligned} \beta' &\mathrel{+}= \gamma'' \cdot \beta \\ \beta'' &\mathrel{+}= \gamma' \cdot \beta R(Z \overset{*}{\Rightarrow} Y) \end{aligned}$$

The first summation is carried out once for each state $j : {}_kX \to \lambda.Z\mu$, whereas the second summation is applied for each choice of $Z$, but only if $X \to \lambda Z\mu$ is not itself a unit production, i.e., $\lambda\mu \neq \epsilon$.

*Rationale.* This increments $\beta''$ the equivalent of $R(Z \overset{*}{\Rightarrow} Y)$ times, accounting for the infinity of surroundings in which $Y$ can occur if it can be derived through cyclic productions. Note that the computation of $\beta'$ is unchanged, since $\gamma''$ already includes an infinity of cyclically generated subtrees for $Y$, where appropriate.

## 5.3 Parsing bracketed inputs

The estimation procedure described above (and EM-based estimators in general) are only guaranteed to find locally optimal parameter estimates. Unfortunately, it seems that in the case of unconstrained SCFG estimation local maxima present a very real problem, and make success dependent on chance and initial conditions (Lari & Young 1990). Pereira & Schabes (1992) showed that partially bracketed input samples can alleviate the problem in certain cases. The bracketing information constrains the parse of the inputs, and therefore the parameter estimates, steering it clear from some of the suboptimal solutions that could otherwise be found.

A second advantage of bracketed inputs is that they potentially allow more efficient processing, since the space of potential derivations (or equivalently, Earley paths) is reduced. It is therefore interesting to see how any given parser can incorporate partial bracketing information. This is typically not a big problem, but in the case of Earley's algorithm there is a particularly simple and elegant solution.

Consider again the grammar

$$\begin{aligned} S &\rightarrow a \quad [p] \\ S &\rightarrow SS \quad [q] \end{aligned}$$

A partially bracketed input for this grammar would be $a(aa)a$. The parentheses indicate phrase boundaries that any candidate parse has to be consistent with, e.g., there cannot be a parse that has a constituent spanning the first and second $a$, or the third and fourth. The supplied bracketing can be nested, of course, and need not be complete, i.e., within a bracketing there are still potentially several ways of parsing a substring.

The Earley parser can deal efficiently with partial bracketing information as follows. A partially bracketed input is processed as usual, left-to-right. When a bracketed portion is encountered, the parser invokes itself recursively on the substring delimited by the pair of parentheses encountered. More precisely:

- The recursive parser instance gets to see only the substring as input.

- Its chart is disjoint from the one used by the parent instance. It cannot use states from the parent chart, except those explicitly passed to it (see below). Conversely, when finished, the parent has access only to those states returned explicitly by the child instance.[13] (The first restriction prevents parsing of constituents that cross the left phrase boundary, while the second restriction prevents a violation of the right phrase boundary.)

- The chart of the child is initialized with all incomplete states from the parent's state set at the start position of the substring.

- The child returns to the parent all (and only) the complete states from its last state set. The parent adds the returned states to the state set at the position immediately following the end of the substring, using it as the input for its own completion procedure.

- Thus the recursive parser invocation and the following completion step replaces the usual prediction-scanning-completion cycle for the entire bracketed substring. After the child returns, the parent continues processing regular input symbols, or other bracketed substrings.

- Needless to say, the child parser instance may itself call on recursive instances to deal with nested bracketings.

This recursion scheme is efficient in that it never explicitly *rejects* a parse that would be inconsistent with the bracketing. Instead it *only considers* those parses that are consistent with the bracketing, while continuing to make use of top-down information like a standard Earley parser.

Processing bracketed strings requires no modification to the computation of probabilities. Probabilities are passed between parent and child as part of states, and processed as before. The recursive control structure simply constrains the set of Earley paths considered by the parser, thereby affecting the probabilities indirectly. For example, ambiguous strings may end up with lower inner probabilities because some derivations are inconsistent with the bracketing.

---

[13]This does not preclude using a shared chart at the implementation level, of course, as long as the above protocol is adhered to.

Only the forward pass is directly affected by the bracketing. Both the Viterbi procedure (Section 5.1) and the reverse completion pass (Section 5.2) only examine the states already in the chart. They are therefore automatically constrained by the bracketing.

**Complexity**  To assess the complexity benefit of bracketing we can extend the analysis of Section 4.8, making use of the recursive structure of the algorithm.

In the standard parsing scheme, the time complexity is $O(l^3)$ for an input of length $l$. Hence, in the recursive scheme each bracketed substring takes time $O(r^3)$, where $r$ is the number of constituents in the substring (which may be either input symbols or nested constituents). The total number of bracketings in a given input string is $O(l)$. If $R$ is an upper bound on $r$ the total time is therefore $O(lR^3)$.

In a *fully bracketed* input string each grammar rule used in the derivation is reflected in a corresponding bracketing. Hence, $r$ is bounded by the maximal production length of the grammar, and the time complexity is simply $O(l)$.

## 5.4  Robust parsing

In many applications ungrammatical input has to be dealt with in some way. Traditionally it was seen as a drawback of top-down parsing algorithms such as Earley's that they sacrifice "robustness," i.e., the ability to find partial parses in an ungrammatical input, for the efficiency gained from top-down prediction (Magerman & Weir 1992).

One approach to the problem is to build robustness into the grammar itself. In the simplest case one could add top-level productions

$$S \rightarrow XS$$
$$\rightarrow \epsilon$$

where $X$ can expand to any nonterminal, including an "unknown word" category. This grammar will cause the Earley parser to find all partial parses of substrings, effectively behaving like a bottom-up parser constructing the chart in left-to-right fashion. More refined variations are possible: the top-level productions could be used to model which phrasal categories (sentence fragments) can likely follow each other. This probabilistic information can then be used in a pruning version of the Earley parser (Section 6.2) to arrive at a compromise between robust and expectation-driven parsing.

An alternative method for making Earley parsing more robust is to modify the parser itself so as to accept arbitrary input and find all or a chosen subset of possible substring parses. Below we present such a simple extension to Earley's algorithm (probabilistic or not). In the probabilistic version, it will also produce the likelihoods of those partial parses. The potential advantage over the grammar modifying approach is that it can be modified to make use of various criteria for which partial parses to allow at runtime.

The extension for robust parsing does not require any changes to the way the Earley parser operates on the chart, only that the chart be "seeded" with some extra states before starting. The computation performed as a result of this modification will be essentially equivalent to that of a CYK bottom-up parser, but with the advantage that a single parsing engine can be used for both standard and robust parsing.

### 5.4.1  Seeding the chart

In standard Earley parsing the parser expects to find exactly one instance of an $S$ nonterminal generating the entire input. This expectation is reflected by the fact that the chart is initialized with the dummy start state

$$0 :_0 \quad \to .S \quad .$$

For robust parsing, we want to identify all nonterminals that can possibly generate any substring of the input. This can be accomplished by also placing dummy states

$$k :_k \quad \to .X \quad ,$$

for all positions $k$ and nonterminals $X$, in the Earley chart prior to the start of normal operation. (In practice, dummy states need to be added only for those nonterminals $X$ whose expansion can start with the current input symbol. This technique is discussed in Appendix B.3.2.)

The immediate effect of these extra states is that more predictions will be generated, from which more completions follow, etc. After finishing the processing of the $j$th state set, the chart will contain states

$$j :_k \quad \to X.$$

indicating that nonterminal $X$ generates the substring $x_{k \ldots j-1}$.

Table 3(a) illustrates the robust parsing process using the example grammar from Table 1 (p. 5).

Probabilities in the extra states are handled as follows. The initial dummy states $_k \to .X$ are initialized with a forward probability of zero. This will ensure that the forward probabilities of all extra states remain at zero and don't interfere with the computation of prefix probabilities from the regular Earley states.

Inner probabilities on dummy states are initialized to unity just as for the $S$ start state, and processed in the usual way. The inner probabilities for the each substring/nonterminal pair can then be read off of the complete dummy states.

Viterbi probabilities and Viterbi back-pointers can also be processed unchanged. Applying the **Viterbi-parse** procedure from Section 5.1 to the complete dummy states yields Viterbi parses for all substring/nonterminal pairs.

### 5.4.2 Assembling partial parses

Instead of consulting the chart for individual substring/nonterminal pairs it may be useful to obtain a list of all complete partial parses of the input. A *complete partial parse* is a sequence of nonterminals that together generate the input. For example, using the grammar in Table 1, the input *a circle touches above a square* has the complete partial parses 'NP VT PP' and 'Det N VT P NP', among others. The input is grammatical exactly if $S$ is among the complete partial parses.

First note that there may not exist a complete partial parse if the input contains unknown symbols. As a preprocessing step, or on-line during parsing, one may have to create new preterminals to account for such new input symbols.

The Earley algorithm can be minimally extended to also generate the list of all partial parses. What is needed is some device that assembles abutting nonterminals from partial parses left-to-right. This work can be carried out as a by-product of the normal completion process using the concept of a *wildcard state*. A wildcard state is a special kind of dummy state in which the RHS can have any number of nonterminals to the left of the dot, and a *wildcard* ? to the right of the dot:

$$i :_k \quad \to \lambda.?$$

As usual, such a state means that the nonterminals in $\lambda$ have generated the substring $x_{k \ldots i-1}$. The wildcard indicates that any continuation of the nonterminal sequence is allowed.

The wildcard semantics are taken into account during prediction and completion. A wildcard state generates predictions for all nonterminals, thus having the same effect as the nonterminal-specific dummy states in the previous section. During completion, a wildcard state combines with

(a)

```
    STATE SET 0                              STATE SET 4
         0         --> .S                      scanned "below" ...
    predicted ...                                   3 P        --> below .
         0 S       --> .NP VP                 completed ...
         0 NP      --> .DET N                      3          --> P .
    STATE SET 1                                MAX  3 P        --> below .
     scanned "a" ...                                3 PP       --> P .NP
         0 DET     --> a .                    predicted ...
    completed ...                                   4 S        --> .NP VP
         0         --> DET .                       4 NP       --> .DET N
    MAX  0 DET     --> a .                    STATE SET 5
         0 NP      --> DET .N                  scanned "a" ...
    predicted ...                                   4 DET      --> a .
    STATE SET 2                                completed ...
     scanned "circle" ...                           4          --> DET .
         1 N       --> circle .               MAX  4 DET      --> a .
    completed ...                                   4 NP       --> DET .N
         0         --> NP .                   predicted ...
         1         --> N .                    STATE SET 6
    MAX  0 NP      --> DET N .                 scanned "square" ...
         1 N       --> circle .                     5 N        --> square .
         0 S       --> NP .VP                 completed ...
    predicted ...                                   3          --> PP .
         2 VP      --> .VI PP                       4          --> NP .
         2 VP      --> .VT NP                 MAX  3 PP       --> P NP .
    STATE SET 3                                     5          --> N .
     scanned "touches" ...                          4 NP       --> DET N .
         2 VT      --> touches .                    5 N        --> square .
    completed ...                                   4 S        --> NP .VP
         2         --> VT .
    MAX  2 VT      --> touches .
         2 VP      --> VT .NP
    predicted ...
         3 PP      --> .P NP
```

(b)

```
    STATE SET 0                              STATE SET 4
         0         --> .?                           0          --> NP VT P .?
    STATE SET 1                                     0          --> DET N VT P .?
         0         --> DET .?                 STATE SET 5
    STATE SET 2                                     0          --> NP VT P DET .?
         0         --> NP .?                        0          --> DET N VT P DET .?
         0         --> DET N .?               STATE SET 6
    STATE SET 3                                     0          --> NP VT PP .?
         0         --> NP VT .?                     0          --> DET N VT PP .?
         0         --> DET N VT .?                  0          --> NP VT P NP .?
                                                    0          --> DET N VT P NP .?
                                                    0          --> NP VT P DET N .?
                                                    0          --> DET N VT P DET N .?
```

Table 3: Robust parsing using the simple grammar from Table 1. (a) State sets generated from parsing the ungrammatical string *a circle touches above a square*. Dummy states (those with empty LHS) represent partial parses. States representing "maximal" partial parses are marked with **MAX**. Predictions that don't lead to completions have been omitted to save space. (b) Trace of wildcard state completions resulting in a list of complete partial parses for this input.

all complete states to yield new wildcard states:

$$\left. \begin{array}{l} i: \quad {}_jY \to \mu. \\ j: \quad {}_k \to \lambda.? \end{array} \right\} \quad \Longrightarrow \quad i: \quad {}_k \to \lambda Y.?$$

for all $Y$. (That is, the complete nonterminal $Y$ is inserted before the dot and the wildcard is retained following the dot to allow further completions.) This is implemented by a trivial modification to the standard completion step. Inner probabilities, Viterbi probabilities and Viterbi back-pointers are processed as usual.

The net effect of processing wildcard states is that all complete partial parses can be read off the final state set in the chart as the right-hand sides of wildcard states (after discarding the wildcard itself). The inner probability of a wildcard state reflects the combined likelihood of the partial parse for the given input. The Viterbi probability of a wildcard state is the joint maximum achieved by the most likely parses for each of the substrings. Different derivations from the same complete partial parse may split the input string at different places. The **Viterbi-parse** procedure when applied to a wildcard state will recover the most likely such split.

Table 3(b) shows a trace of wildcard state completions used in enumerating the partial parses for the example given earlier.

The total number of complete partial parses can be exponential in the length of the input. It may therefore be desirable to compute only a subset of them, applying some application-specific filter criterion. One such criterion is that one is only interested in "maximal" complete partial parses. A complete partial parse is called *maximal* if it has no subsequence of nonterminals that can be replaced by another nonterminal so as to yield another complete partial parse. For example, in the case of *a circle touches above a square*, the only maximal parse is 'NP VT PP'.

It turns out that a filter for maximal partial parses is easy to implement in the Earley framework. Maximal parses contain only nonterminals that are not themselves part of a larger partial parse. Therefore, during completion, we can mark all states that contributed to a larger constituent, and later identify the unmarked states as the ones corresponding to maximal parses. (The chart in Table 3(a) has all maximal states labeled with `MAX`.) When completing wildcard states we simply skip all completions due to non-maximal states. The list of complete partial parses obtained from the chart will then contain precisely the maximal ones.

## 6 Discussion

### 6.1 Relation to finite-state models

Throughout the exposition of the Earley algorithm and its probabilistic extension we have been alluding, in concepts and terminology, to the algorithms used with probabilistic finite-state models, in particular Hidden Markov Models (Rabiner & Juang 1986). Many concepts carry over, if suitably generalized, most notably that of forward probabilities. Prefix probabilities can be computed from forward probabilities by the Earley parser just as in HMMs because Earley states summarize past history in much the same way as the states in a finite-state model. There are important differences, however. The number of states in an HMM remains fixed, whereas the number of possible Earley states grows linearly with the length of the input (due to the start index).

Incidentally, the HMM concept of "backward probabilities" has no useful analog in Earley parsing. It is tempting to define $\beta_i(s)$ as the conditional probability that the generator produces the remaining string given that it is currently in state $s$. Alas, this would be an ill-defined quantity since the generation of a suffix depends (via completion) on more than just the current state.

The solution found by Baker (1979), adopted here in modified form, is to use outer probabilities instead of "backward" probabilities. Outer probabilities follow the hierarchical structure of a derivation, rather than the sequential structure imposed by left-to-right processing. Fortunately, outer probability computation is just as well supported by the Earley chart as forward and inner probabilities.[14]

## 6.2 Online pruning

In finite-state parsing (especially speech decoding) one often makes use of the forward probabilities for *pruning* partial parses before having seen the entire input. Pruning is formally straightforward in Earley parsers: in each state set, rank states according to their $\alpha$ values, then remove those states with small probabilities compared to the current best candidate, or simply those whose rank exceed a given limit. Notice this will not only omit certain parses, but will also underestimate the forward and inner probabilities of the derivations that remain. Pruning procedures have to be evaluated empirically since they invariably sacrifice completeness and, in the case of the Viterbi algorithm, optimality of the result.

While Earley-based on-line pruning awaits further study, there is reason to believe the Earley framework has inherent advantages over strategies based only on bottom-up information (including so-called "over-the-top" parsers). Context-free forward probabilities include *all* available probabilistic information (subject to assumptions implicit in the SCFG formalism) available from an input prefix, whereas the usual inside probabilities do not take into account the nonterminal prior probabilities that result from the top-down relation to the start state. Using top-down constraints does not necessarily mean sacrificing robustness, as discussed in Section 5.4. On the contrary, by using Earley-style parsing with a set of carefully designed and estimated "fault tolerant" top-level productions, it should be possible to use probabilities to better advantage in robust parsing. This approach is a subject of ongoing work, in the context of tight-coupling SCFGs with speech decoders (Jurafsky *et al.* 1994a).

## 6.3 Relation to probabilistic LR parsing

One of the major alternative context-free parsing paradigms besides Earley's algorithm is *LR parsing* (Aho & Ullman 1972). A comparison of the two approaches, both in their probabilistic and non-probabilistic aspects, is interesting and provides useful insights. The following remarks assume familiarity with both approaches. We sketch the fundamental relations, as well as the important tradeoffs between the two frameworks.[15]

Like an Earley parser, LR parsing uses dotted productions, called *items*, to keep track of the progress of derivations as the input is processed. The start indices are not part of LR items: we may therefore use the term "item" to refer to both LR items and Earley states without start indices. An Earley parser constructs sets of possible items on the fly, by following all possible partial derivations. An LR parser, on the other hand, has access to a complete list of *sets of possible items* computed beforehand, and at runtime simply follows transitions between these sets. The item sets are known as the "states" of the LR parser.[16] A grammar is suitable for LR parsing if these transitions can be performed deterministically by considering only the next input and the contents of a shift-reduce stack. *Generalized LR parsing* is an extension that allows parallel tracking of multiple state transitions and stack actions by using a graph-structured stack (Tomita 1986).

*Probabilistic LR parsing* (Wright 1990) is based on LR items augmented with certain conditional probabilities. Specifically, the probability $p$ associated with an LR item $X \rightarrow \lambda.\mu$ is, in our

---

[14] The closes thing to a HMM backward probability is probably the *suffix probability* $P(S \stackrel{*}{\Rightarrow}_R x)$.

[15] Like Earley parsers, LR parsers can be built using various amounts of *lookahead* to make the operation of the parser (more) deterministic, and hence more efficient. Only the case of zero-lookahead, LR(0), is considered here; the correspondence between LR($k$) parsers and $k$-lookahead Earley parsers is discussed in the literature (Earley 1970; Aho & Ullman 1972).

[16] Once again, it is helpful to compare this to a closely related finite-state concept: the states of the LR parser correspond to sets of Earley states, similar to the way the states of a deterministic FSA correspond to sets of states of an equivalent non-deterministic FSA under the standard subset construction.

terminology, a normalized forward probability:

$$p = \frac{\alpha_i(X \to \lambda.\mu)}{P(S \stackrel{*}{\Rightarrow}_L x_{0...i-1})} \quad ,$$

where the denominator is the probability of the current prefix.[17] LR item probabilities, are thus conditioned forward probabilities, and can be used to compute conditional probabilities of next words: $P(x_i|x_{0...i-1})$ is the sum of the $p$'s of all items having $x_i$ to the right of the dot (extra work is required if the item corresponds to a "reduce" state, i.e., if the dot is in final position).

Notice that the definition of $p$ is independent of $i$ as well as the start index of the corresponding Earley state. Therefore, to ensure that item probabilities are correct independent of input position, item sets would have to be constructed so that their probabilities are unique within each set. However, this may be impossible given that the probabilities can take on infinitely many values and in general depend on the history of the parse. The solution used by Wright (1990) is to collapse items whose probabilities are within a small tolerance $\epsilon$ and are otherwise identical. The same threshold is used to simplify a number of other technical problems, e.g., left-corner probabilities are computed by iterated prediction, until the resulting changes in probabilities are smaller than $\epsilon$. Subject to these approximations, then, a probabilistic LR parser can compute prefix probabilities by multiplying successive conditional probabilities for the words it sees.[18]

As an alternative to the computation of LR transition probabilities from a given SCFG, one might instead estimate such probabilities directly from traces of parses on a training corpus. Due to the imprecise relationship between LR probabilities and SCFG probabilities it is not clear if the model thus estimated corresponds to any particular SCFG in the usual sense.

Briscoe & Carroll (1993) turn this incongruity into an advantage by using the LR parser as a probabilistic model in its own right, and show how LR probabilities can be extended to capture non-context-free contingencies. The problem of capturing more complex distributional constraints in natural language is clearly important, but well beyond the scope of this paper. We simply remark that it should be possible to define "interesting" non-standard probabilities in terms of Earley parser actions so as to better model non-context-free phenomena.

Apart from such considerations, the choice between LR methods and Earley parsing is a typical space-time tradeoff. Even though an Earley parser runs with the same linear time and space complexity as an LR parser on grammars of the appropriate LR class, the constant factors involved will be much in favor of the LR parser as almost all the work has already been compiled into its transition and action table. However, the size of LR parser tables can be exponential in the size of the grammar (due to the number of potential item subsets). Furthermore, if the generalized LR method is used for dealing with non-deterministic grammars (Tomita 1986) the runtime on arbitrary inputs may also grow exponentially. The bottom line is that each application's needs have to be evaluated against the pros and cons of both approaches to find the best solution. From a theoretical point of view, the Earley approach has the inherent appeal of being the more general (and exact) solution to the computation of the various SCFG probabilities.

### 6.4 Other related work

The literature on Earley-based probabilistic parsers is sparse, presumably because of the precedent set by the Inside/Outside algorithm, which is more naturally formulated as a bottom-up algorithm.

Both Nakagawa (1987) and Päseler (1988) use a non-probabilistic Earley parser augmented with "word match" scoring. Though not truly probabilistic, these algorithms are similar to the Viterbi version described here, in that they find a parse that optimizes the accumulated matching scores (without regard to rule probabilities). Prediction and completion loops do not come into play since no precise inner or forward probabilities are computed.

---

[17] The identity of this expression with the item probabilities of Wright (1990) can be proved by induction on the steps performed to compute the $p$'s, as shown in Appendix C.

[18] It is not clear what the numerical properties of this approximation are, e.g., how the errors will accumulate over longer parses.

Magerman & Marcus (1991) are interested primarily in scoring functions to guide a parser efficiently to the most promising parses. Earley-style top-down prediction is used only to suggest worthwhile parses, not to compute precise probabilities, which they argue would be an inappropriate metric for natural language parsing.

Casacuberta & Vidal (1988) exhibit an Earley parser that processes weighted (not necessarily probabilistic) CFGs and performs a computation that is isomorphic to that of inside probabilities shown here. Schabes (1991) adds both inner and outer probabilities to Earley's algorithm, with the purpose of obtaining a generalized estimation algorithm for SCFGs. Both of these approaches are restricted to grammars without unbounded ambiguities, which can arise from unit or null productions.

Dan Jurafsky (personal communication) wrote an Earley parser for the Berkeley Restaurant Project (BeRP) speech understanding system that originally computed forward probabilities for restricted grammars (without left-corner or unit production recursion). The parser now uses the method described here to provide exact SCFG prefix and next-word probabilities to a tightly-coupled speech decoder (Jurafsky *et al.* 1994a).

An essential idea in the probabilistic formulation of Earley's algorithm is the collapsing of recursive predictions and unit completion chains, replacing both with lookups in precomputed matrices. This idea arises in our formulation out of the need to compute probability sums given as infinite series. Graham *et al.* (1980) use a non-probabilistic version of the same technique to create a highly optimized Earley-like parser for general CFGs that implements prediction and completion by operations on Boolean matrices.[19]

The matrix inversion method for dealing with left-recursive prediction is borrowed from the LRI algorithm of Jelinek & Lafferty (1991) for computing prefix probabilities for SCFGs in CNF.[20] We then use that idea a second time to deal with the similar recursion arising from unit productions in the completion step. We suspect, but have not proved, that the Earley computation of forward probabilities when applied to a CNF grammar performs a computation that is isomorphic to that of the LRI algorithm. In any case, we believe that the parser-oriented view afforded by the Earley framework makes for a very intuitive solution to the prefix probability problem, with the added advantage that it is not restricted to CNF grammars.

Kupiec (1992) has proposed a version of the Inside/Outside algorithm that allows it to operate on non-CNF grammars. Interestingly, Kupiec's algorithm is also based on a generalization of finite-state models, namely, Recursive Transition Networks (RTNs). Probabilistic RTNs are essentially HMMs that allow nonterminals as output symbols. Also, the dotted productions appearing in Earley states are of course exactly equivalent to the states in an RTN derived from a CFG.

## 6.5 A simple typology of SCFG algorithms

The various known algorithms for probabilistic CFGs share many similarities, and vary along similar dimensions. One such dimension is whether the quantities entered into the parser chart are defined in a bottom-up (CYK) fashion, or whether left-to-right constraints are an inherent part of their definition.[21]

Another point of variation is the "sparseness" trade-off. If we are given a set of nonterminals and wanted to list all possible CFG rules involving those nonterminals, the list would be infinite due to the arbitrary length of the right-hand sides of productions. This is a problem, for example, when training a CFG starting with complete ignorance about the structure of the rules.

---

[19] This connection to the GHR algorithm was pointed out by Fernando Pereira. Exploration of this link then lead to the extension of our algorithm to handle ϵ-productions, as described in Section 4.7.

[20] Their method uses the transitive (but not reflexive) closure over the left-corner relation $P_L$, for which they chose the symbol $Q_L$. We chose the symbol $R_L$ in this paper to point to this difference.

[21] Of course a CYK-style parser can operate left-to-right, right-to-left, or otherwise by reordering the computation of chart entries.

|  | Full CNF | Sparse CFG |
|---|---|---|
| Bottom-up | Inside/outside (Baker 1979) | Stochastic RTNs (Kupiec 1992) |
| Left-to-right | LRI (Jelinek & Lafferty 1991) | Probabilistic Earley |

Table 4: Tentative typology of SCFG algorithms according to prevailing directionality and sparseness of the CFG.

A workaround is to restrict the rule format somehow, usually to CNF, and then list all possible productions. Algorithms that assume CNF are usually formulated in terms of such a fully parameterized grammar where all triples $X, Y, Z$ form a possible rule $X \rightarrow YZ$ with non-zero probability, although in many cases they may be specialized to handle sparse grammars efficiently.

At the other extreme we have algorithms with accept unrestricted CFG productions and are therefore meant for sparse grammars, where almost all (in the set theoretic sense) possible productions have probability zero. It appears that these algorithms tend to be more naturally formulated in terms of a stochastic process, as opposed to static specifications of string probabilities.

To illustrate these points, the algorithms discussed in this section have been arranged in the grid depicted in Table 4.

# 7 Conclusions

We have presented an Earley-based parser for stochastic context-free grammars that is appealing for its combination of advantages over existing methods. Earley's control structure lets the algorithm run with best-known complexity on a number of grammar subclasses, and no worse than standard bottom-up probabilistic chart parsers on general SCFGs and fully parameterized CNF grammars.

Unlike bottom-up parsers it also computes accurate prefix probabilities incrementally while scanning its input, along with the usual substring (inside) probabilities. The chart constructed during parsing supports both Viterbi parse extraction and Baum-Welch type rule probability estimation by way of a backward pass over the parser chart. If the input comes with (partial) bracketing to indicate phrase structure this information can be easily incorporated to restrict the allowable parses. A simple extension of the Earley chart allows finding partial parses of ungrammatical input.

The computation of probabilities is conceptually simple, and follows directly Earley's parsing framework, while drawing heavily on the analogy to finite-state language models. It does not require rewriting the grammar into normal form. Thus, the present algorithm fills a gap in the existing array of algorithms for SCFGs, efficiently combining the functionalities and advantages of several previous approaches.

# Acknowledgments

Thanks are due Dan Jurafsky and Steve Omohundro for extensive discussions on the topics in this paper, and Fernando Pereira for helpful advice and pointers. Jerry Feldman, Terry Regier, Jonathan Segal, Kevin Thompson and four anonymous reviewers provided valuable comments for improving the presentation.

# A   Appendix: Existence of $R_L$ and $R_U$

In Section 4.5 we defined the probabilistic left-corner and unit-production matrices $R_L$ and $R_U$, respectively, to collapse recursions in the prediction and completion steps. It was shown how these matrices could be obtained as the result of matrix inversions. In this appendix we give a proof that the existence of these inverses is assured if the grammar is well-defined in the following three senses. The terminology used here is taken from Booth & Thompson (1973).

**Definition 9**  For a SCFG $G$ over an alphabet $\Sigma$, with start symbol $S$, we say that[22]

a) $G$ is *proper* iff for all nonterminals $X$ the rule probabilities sum to unity, i.e.,

$$\sum_{\lambda:(X\to\lambda)\in G} P(X \to \lambda) = 1 \quad.$$

b) $G$ is *consistent* iff it defines a probability distribution over finite strings, i.e.,

$$\sum_{x\in\Sigma^*} P(S \stackrel{*}{\Rightarrow} x) = 1 \quad,$$

where $P(S \stackrel{*}{\Rightarrow} x)$ is induced by the rule probabilities according to Definition 1(a).

c) $G$ has *no useless nonterminals* iff all nonterminals $X$ appear in at least one derivation of some string $x \in \Sigma^*$ with non-zero probability, i.e., $P(S \stackrel{*}{\Rightarrow} \lambda X \mu \stackrel{*}{\Rightarrow} x) > 0$.

It is useful to translate consistency into "process" terms. We can view an SCFG as a stochastic string-rewriting process, in which each step consists of simultaneously replacing all nonterminals in a sentential form with the right-hand sides of productions, randomly drawn according to the rule probabilities. Booth & Thompson (1973) show that the grammar is consistent if and only if the probability that stochastic rewriting of the start symbol $S$ leaves nonterminals remaining after $n$ steps, goes to 0 as $n \to \infty$. More loosely speaking, rewriting $S$ has to terminate after a finite number of steps with probability 1, or else the grammar is inconsistent.

We observe that the same property holds not only for $S$, but for all nonterminals, if the grammar has no useless terminals. If any nonterminal $X$ admitted infinite derivations with non-zero probability, then $S$ itself would have such derivations, since by assumption $X$ is reachable from $S$ with non-zero probability.

To prove the existence of $R_L$ and $R_U$, it is sufficient to show that the corresponding geometric series converge:

$$\begin{aligned} R_L &= I + P_L + P_L^2 + \ldots = (I - P_L)^{-1} \\ R_U &= I + P_U + P_U^2 + \ldots = (I - P_U)^{-1} \quad. \end{aligned}$$

**Lemma 5**  If $G$ is a proper, consistent SCFG without useless nonterminals, then the powers $P_L^n$ of the left-corner relation, and $P_U^n$ of the unit-production relation, converge to zero as $n \to \infty$.

*Proof.*  Entry $(X, Y)$ in the left-corner matrix $P_L$ is the probability of generating $Y$ as the immediately succeeding left-corner below $X$. Similarly, entry $(X, Y)$ in the $n$th power $P_L^n$ is the probability of generating $Y$ as the left-corner of $X$ with $n - 1$ intermediate nonterminals. Certainly $P_L^n(X, Y)$ is bounded above by the probability that the entire derivation starting at $X$ terminates after $n$ steps, since a derivation couldn't terminate without expanding the left-most symbol to a terminal (as opposed to a nonterminal). But that probability tends to 0 as $n \to \infty$, and hence so must each entry in $P_L^n$.

---

[22]Unfortunately, the terminology used in the literature is not uniform. For example, Jelinek & Lafferty (1991) use the term "proper" to mean (c), and "well-defined" for (b). They also state mistakenly that (a) and (c) together are a sufficient condition for (b). Booth & Thompson (1973) show that one can write a SCFG that satisfies (a) and (c) but generates derivations that do not terminate with probability 1, and give necessary and sufficient conditions for (b).

For the unit-production matrix $P_U$ a similar argument applies, since the length of a derivation is at least as long as it takes to terminate any initial unit-production chain.

**Lemma 6** If $G$ is a proper, consistent SCFG without useless nonterminals, then the series for $R_L$ and $R_U$ as defined above converge to finite, non-negative values.

*Proof.* $P_L^n$ converging to 0 implies that the magnitude of $P_L$'s largest eigenvalue (its spectral radius) is $< 1$, which in turn implies that the series $\sum_{i=0}^{\infty} P_L^i$ converges (similarly for $P_U$). The elements of $R_L$ and $R_U$ are non-negative since they are the result of adding and multiplying among the non-negative elements of $P_L$ and $P_U$, respectively.

Interestingly, a SCFG may be inconsistent and still have converging left-corner and/or unit-production matrices, i.e., consistency is a stronger constraint. For example

$$\begin{aligned} S &\to a & [p] \\ S &\to SS & [q] \end{aligned}$$

is inconsistent for any choice of $q \geq \frac{1}{2}$, but the left-corner relation (a single number in this case) is well-defined for all $q < 1$, namely $(1-q)^{-1} = p^{-1}$. In this case the left fringe of the derivation is guaranteed to result in a terminal after finitely many steps, but the derivation as a whole may never terminate.

# B  Appendix: Implementation Notes

This appendix discusses some of the experiences gained from implementing the probabilistic Earley parser.

## B.1  Prediction

Due to the collapsing of transitive predictions, this step can be implemented in a very efficient and straightforward manner. As explained in Section 4.5, one has to perform a single pass over the current state set, identifying all nonterminals $Z$ occurring to the right of dots, and add states corresponding to all productions $Y \to \nu$ that are reachable through the left-corner relation $Z \stackrel{*}{\Rightarrow}_L Y$. As indicated in equation (3), contributions to the forward probabilities of new states have to be summed when several paths lead to the same state. However, the summation in equation (3) can be optimized if the $\alpha$ values for all old states with the same nonterminal $Z$ are summed first, and then multiplied by $R(Z \stackrel{*}{\Rightarrow}_L Y)$. These quantities are then summed over all nonterminals $Z$, and the result is once multiplied by the rule probability $P(Y \to \nu)$ to give the forward probability for the predicted state.

## B.2  Completion

Unlike prediction, the completion step still involves iteration. Each complete state derived by completion can potentially feed other completions. An important detail here is to ensure that all contributions to a state's $\alpha$ and $\gamma$ are summed before proceeding with using that state as input to further completion steps.

One approach to this problem is to insert complete states into a prioritized queue. The queue orders states by their start indices, highest first. This is because states corresponding to later expansions always have to be completed first before they can lead to the completion of expansions earlier on in the derivation. For each start index, the entries are managed as a first-in-first-out queue, ensuring that the dependency graph formed by the states is traversed in breadth-first order.

The completion pass can now be implemented as follows. Initially, all complete states from the previous scanning step are inserted in the queue. States are then removed from the front of the queue, and used to complete other states. Among the new states thus produced, complete ones are again added to the queue. The process iterates until no more states remain in the queue. Because the computation of probabilities already includes chains of unit productions, states derived from such productions need not be queued, which also ensures that the iteration terminates.

A similar queuing scheme, with the start index order reversed, can be used for the reverse completion step needed in the computation of outer probabilities (Section 5.2).

## B.3 Efficient parsing with large sparse grammars

During work with a moderate-sized, application-specific natural language grammar taken from the BeRP speech system (Jurafsky *et al.* 1994b) we had opportunity to optimize our implementation of the algorithm. Below we relate some of the lessons learned in the process.

### B.3.1 Speeding up matrix inversions

Both prediction and completion steps make use of a matrix $R$ defined as a geometric series derived from a matrix $P$,

$$R = I + P + P^2 + \ldots = (I - P)^{-1} \quad .$$

Both $P$ and $R$ are indexed by the nonterminals in the grammar. The matrix $P$ is derived from the SCFG rules and probabilities (either the left-corner relation or the unit-production relation).

For an application using a fixed grammar the time taken by the precomputation of left-corner and unit-production matrices may not be crucial, since it occurs off-line. There are cases, however, when that cost should be minimal, e.g., when rule probabilities are iteratively reestimated.

Even if the matrix $P$ is sparse, the matrix inversion can be prohibitive for large numbers of nonterminals $n$. Empirically, matrices of rank $n$ with a bounded number $p$ of non-zero entries in each row (i.e., $p$ is independent of $n$) can be inverted in time $O(n^2)$, whereas a full matrix of size $n \times n$ would require time $O(n^3)$.

In many cases the grammar has a relatively small number of nonterminals that have productions involving other nonterminals in a left-corner (or the RHS of a unit-production). Only those nonterminals can have non-zero contributions to the higher powers of the matrix $P$. This fact can be used to substantially reduce the cost of the matrix inversion needed to compute $R$.

Let $P'$ be a subset of the entries of $P$, namely, only those elements indexed by nonterminals that have a non-empty row in $P$. For example, for the left-corner computation, $P'$ is obtained from $P$ by deleting all rows and columns indexed by nonterminals that do not have productions starting with nonterminals. Let $I'$ be the identity matrix over the same set of nonterminals as $P'$. Then $R$ can be computed as

$$\begin{aligned} R &= I + (I + P + P^2 + \ldots)P \\ &= I + (I' + P' + P'^2 + \ldots) \star P \\ &= I + (I' - P')^{-1} \star P \\ &= I + R' \star P \quad . \end{aligned}$$

Here $R'$ is the inverse of $I' - P'$, and $\star$ denotes a matrix multiplication in which the left operand is first augmented with zero elements to match the dimensions of the right operand, $P$.

The speedups obtained with this technique can be substantial. For a grammar with 789 nonterminals, of which only 132 have nonterminal productions, the left-corner matrix was computed in 12 seconds (including the final multiply with $P$ and addition of $I$). Inversion of the full matrix $I - P$ took 4 minutes 28 seconds.[23]

### B.3.2 Linking and bottom-up filtering

As discussed in Section 4.8, the worst-case run-time on fully parameterized CNF grammars is dominated by the completion step. However, this is not necessarily true of sparse grammars. Our experiments showed that the computation is dominated by the generation of Earley states during the prediction steps.

It is therefore worthwhile to minimize the total number of predicted states generated by the parser. Since predicted states only affect the derivation if they lead to subsequent scanning we can use the next input symbol to constrain the relevant predictions. To this end, we compute the *extended left-corner relation* $R_{LT}$, indicating which *terminals* can appear as left corners of which nonterminals. $R_{LT}$ is a Boolean matrix with rows indexed by nonterminals and columns indexed by terminals. It can be computed as the product

$$R_{LT} = R_L P_{LT}$$

where $P_{LT}$ has a non-zero entry at $(X, a)$ iff there is a production for nonterminal $X$ that starts with terminal $a$. $R_L$ is the old left-corner relation.

During the prediction step we can ignore incoming states whose RHS nonterminal following the dot cannot have the current input as a left-corner, and then eliminate from the remaining predictions all those whose LHS cannot produce the current input as a left-corner. These filtering steps are very fast as they involve only table lookup.

This technique for speeding up Earley prediction is the exact converse of the "linking" method described by Pereira & Shieber (1987:chapter 6) for improving the efficiency of bottom-up parsers. There, the extended left-corner relation is used for *top-down filtering* the *bottom-up application* of grammar rules. In our case, we use linking to provide *bottom-up filtering* for *top-down application* of productions.

On a test corpus this technique cut the number of generated predictions to almost 1/4 and speeded up parsing by a factor of 3.3. The corpus consisted of 1143 sentence with an average length of 4.65 words. The top-down prediction alone generated 991781 states and parsed at a rate of 590 milliseconds per sentence. With bottom-up filtered prediction only 262287 states were generated, resulting in 180 milliseconds per sentence.

A trivial optimization often found in Earley parsers is to precompute the entire first prediction step, as it doesn't depend on the input and may eliminate a substantial portion of the total predictions per sentence.[24] We found that with bottom-up filtering this technique lost its edge: scanning the precomputed predicted states turned out to be slower than computing the zeroth state set filtered by the first input.

## C  LR item probabilities as conditional forward probabilities

In Section 6.3 an interpretation of LR item probabilities as defined in Wright (1990:Section 2.1) was given in terms of the forward probabilities used by the Earley parser. Below we give a proof for the correctness of this interpretation. Notice that these are the "ideal" LR probabilities that *should* be attached to items, if it weren't for the identification of items with close probabilities to keep the LR state list finite.

---

[23]These figures are not very meaningful for their absolute values. All measurements were obtained on a Sun SPARCstation 2 with a CommonLisp/CLOS implementation of generic sparse matrices that was not particularly optimized for this task.

[24]The first prediction step accounted for roughly 30% of all predictions on our test corpus.

Let $p(X \to \nu)$ be the probability for LR item $X \to \nu$ (with a dot somewhere in the RHS). We want to show that

$$p(X \to \lambda.\mu) = \frac{\alpha_i(_kX \to \lambda.\mu)}{P(S \stackrel{*}{\Rightarrow}_L x_{0...i-1})} \quad , \tag{5}$$

for any item $X \to \lambda.\mu$, regardless of position $i$ and start index $k$. Note that $i$ is not always equal to the position of the last input symbol processed; a reduce action of the parser effectively resets $i$ to the beginning of the reduced nonterminal.

The computation of LR item sets begins with the initial item $\to .S$, which has $p = 1$ by definition, thereby agreeing with (5).

The first operation for constructing item sets is *closure*, whereby for each item $X \to \lambda.Y\mu$, all items $Y \to .\nu$ corresponding to the available productions $Y \to \nu$ are added to the set. This operation is recursive and corresponds obviously to Earley's prediction step. Also, the way in which $p$ values are propagated follows exactly the way forward probabilities are handled during prediction. (The left-corner relation $R_L$ could be used to compute closure probabilities exactly, but Wright suggests using a truncated recursion instead.) Since closure and prediction are thus isomorphic, and since the prefix relative to the items does not change, (5) also remains valid during this step.

Finally, a successor set $I'$ of kernel items is constructed from an existing closed set $I$ in what corresponds to Earley's scanning or completion. Specifically, for each current item $X \to \lambda.Y\mu \in I$, an item $X \to \lambda Y.\mu$ is placed in $I'$, reachable by scanning a terminal $Y$ or reducing (completing) a nonterminal $Y$. (We let $Y$ stand for either terminal or nonterminal to treat both cases jointly.) The new item probability is computed as

$$p(X \to \lambda Y.\mu) = \frac{p(X \to \lambda.Y\mu)}{\sum_{Z \to \pi.Y\rho \in I} p(Z \to \pi.Y\rho)} \tag{6}$$

This can be understood as scaling the total probability of items matching $Y$ to unity.

By substituting (5) into (6) we get

$$\begin{aligned}
p(X \to \lambda Y.\mu) &= \frac{\frac{\alpha_i(X \to \lambda.Y\mu)}{P(S \stackrel{*}{\Rightarrow}_L x_{0...i-1})}}{\sum_{Z \to \pi.Y\rho \in I} \frac{\alpha_i(Z \to \pi.Y\rho)}{P(S \stackrel{*}{\Rightarrow}_L x_{0...i-1})}} \\
&= \frac{\alpha_i(X \to \lambda.Y\mu)}{\sum_{Z \to \pi.Y\rho \in I} \alpha_i(Z \to \pi.Y\rho)} \\
&= \frac{\alpha_i(X \to \lambda.Y\mu)\gamma_{i'}(Y)}{\sum_{Z \to \pi.Y\rho \in I} \alpha_i(Z \to \pi.Y\rho)\gamma_{i'}(Y)} \\
&= \frac{\alpha_{i'}(X \to \lambda Y.\mu)}{\sum_{Z \to \pi Y.\rho \in I'} \alpha_{i'}(Z \to \pi Y.\rho)} \tag{7} \\
&= \frac{\alpha_{i'}(X \to \lambda Y.\mu)}{P(S \stackrel{*}{\Rightarrow}_L x_{0...i'-1})} \tag{8} \\
&= p(X \to \lambda Y.\mu)
\end{aligned}$$

The position $i'$ is that of the current next input. We have used the abbreviation $\gamma_{i'}(Y)$ for the sum of inner probabilities pertaining to the completed $Y$, i.e.,

$$\gamma_i(Y) = \begin{cases} 1 & \text{if } Y \text{ is terminal} \\ \sum_{Y \to \nu.} \gamma_i(Y \to \nu.) & \text{if } Y \text{ is nonterminal.} \end{cases}$$

Two steps in the derivation above need justification. In (7) we are computing forward probabilities just as in an Earley completion step (see equation (1)). To get (8) we observe that the set $I'$ contains *all* possible kernel items after having processed the prefix $x_{0...i'-1}$ (by definition of the LR parsing method). Hence the sum of $\alpha_{i'}$ represents all possible partial derivations generating the prefix, i.e., $P(S \stackrel{*}{\Rightarrow}_L x_{0...i'-1})$.